\documentclass[lettersize,journal]{IEEEtran}
\usepackage{amsmath,amsfonts}
\usepackage{algorithmicx}
\usepackage{algorithm}
\usepackage{algpseudocode}
\usepackage{array}
\usepackage[caption=false,font=normalsize,labelfont=sf,textfont=sf]{subfig}
\usepackage{textcomp}
\usepackage{stfloats}
\usepackage{booktabs}
\usepackage{url}
\usepackage{verbatim}
\usepackage{tabularx}
\usepackage{graphicx}
\usepackage{cite}
\usepackage{rotating}
\usepackage{multirow}
\usepackage{diagbox}
\usepackage[table]{xcolor}
\usepackage[most]{tcolorbox}
\definecolor{Gray}{gray}{0.9}
\definecolor{MyYellow}{RGB}{243,246,204}
\definecolor{MyGreen}{RGB}{248,246,240}
\definecolor{MyBlue}{RGB}{226,235,251}
\definecolor{LightYellow}{RGB}{243,246,233}
\definecolor{LightGray}{RGB}{243,246,250}
\graphicspath{{images/}}
\newtheorem{theorem}{Theorem}[section]
\newtheorem{assumption}[theorem]{Assumption}
\newtheorem{proposition}[theorem]{Proposition}
\newtheorem{corollary}[theorem]{Corollary}
\newtheorem{definition}[theorem]{Definition}
\begin{document}

\title{When Semantically Consistent Encoding Meets View-Label Heterogeneity Modeling: A Unified Framework for Incomplete Multi-View Multi-Label Learning}

\author{Chengliang Liu,
Bo Li,
Bob Zhang, \IEEEmembership{Senior Member}, \emph{IEEE},
Yanghao Zhou,
Jie Wen, \IEEEmembership{Senior Member}, \emph{IEEE}, \\
Wenwu Wang,~\IEEEmembership{Fellow}, \emph{IEEE}
\thanks{This work is supported by the University of Macau (MYRG-GRG2024-00205-FST); the Science and Technology Development Fund, Macao S.A.R (FDCT) (0028/2023/RIA1); Guangdong Provincial Association for Science and Technology Young Scientific and Technological Talent Cultivation Program (SKXRC2026408); and Guangdong Basic and Applied Basic Research Foundation (2024A1515030213).}
\thanks{Chengliang Liu and Bo Li are with the PAMI Research Group, Department of Artificial Intelligence, University of Macau, Macau. E-mail: liucl1996@163.com; bo\_li\_mail@163.com.}
\thanks{Bob Zhang is with the PAMI Research Group, Department of Artificial Intelligence, and Centre for Artificial General Intelligence, Institute of Artificial Intelligence and Brain Sciences, University of Macau, Macau. E-mail: bobzhang@um.edu.mo.}
\thanks{Yanghao Zhou is with the School of Computer Science and Technology, Beijing Institute of Technology, Beijing 100811, China. E-mail: zhouyh77@bit.edu.cn.}
\thanks{Jie Wen is with the School of Computer Science and Technology, Harbin Institute of Technology, Shenzhen 518055, China. E-mail: jiewen\_pr@126.com.}
\thanks{Wenwu Wang is with the School of Computer Science and Electronic Engineering, University of Surrey, Guildford, GU2 7XH,  UK. E-mail: w.wang@surrey.ac.uk.}
\thanks{Corresponding authors: Bob Zhang and Jie Wen.}
}



\maketitle

\begin{abstract}

Incomplete multi-view multi-label learning requires not only robust semantic aggregation from partially observed views, but also label-aware exploitation of view-specific evidence. Existing approaches usually emphasize either shared representation learning or decision-level fusion. The former improves robustness against missing views, yet tends to compress label-discriminative view-specific cues into a single latent representation. The latter preserves individual view predictions, but often relies on fixed or globally learned fusion weights, ignoring that different labels of different instances may require different views.
To address these limitations, this paper presents V2L, a unified representation-decision framework for incomplete multi-view multi-label classification. On the representation side, V2L constructs semantically consistent variational posteriors from incomplete views through a perturbation-aware encoding mechanism, which provides a stable shared semantic basis. On the decision side, V2L introduces an active view-label relevance modeling strategy that estimates instance-wise and label-wise view contributions, allowing each label prediction to adaptively select useful view-specific evidence. From the perspective of model architecture, these two important strategies are integrated into a unified framework through a hybrid fusion architecture, simultaneously meeting the requirements of cross-view semantic consistency and representational complementarity.
{Extensive experiments under both incomplete and complete settings show that V2L achieves leading performance on five benchmarks. Code is available at: \url{https://github.com/justsmart/V2L}}.
\end{abstract}


\section{Introduction}
\IEEEPARstart{M}{ulti}-view multi-label learning aims to predict multiple semantic labels from heterogeneous observations describing the same target \cite{wen2022survey,yu2021multi,liu2024attention}. It has become increasingly important in applications such as image understanding, multimedia retrieval, medical analysis, and multimodal content reasoning, where different views often capture complementary properties of an object and multiple labels are required to express its semantic richness \cite{liu2025reliable, li2022learning}. In such scenarios, effective learning depends on both exploiting cross-view complementarity and modeling label-aware semantics.

{In practice, however, multi-view multi-label data are rarely fully observed \cite{zhang2024discriminative}. Missing views arise from acquisition failures, cost constraints, privacy restrictions, or occlusion, while missing labels are common due to annotation sparsity and human omission \cite{ma2021expand,zhou2024fair,jiang2025collaborative}. As a result, incomplete multi-view multi-label classification must simultaneously cope with partial observations on the input side and incomplete supervision on the output side. Compared with complete-data settings, this dual incompleteness makes the learning problem substantially more difficult, because the model must infer reliable semantics from corrupted evidence while avoiding overfitting to noisy or weakly informative views \cite{li2022concise, zhang2022effective}.}

\begin{figure}[t!]
\centering
\includegraphics[width=0.95\linewidth]{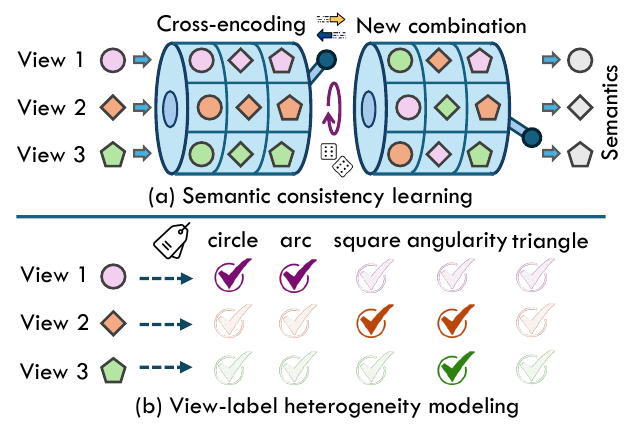} 
\caption{Schematic diagrams of the two core motivations. (a): Cross-view encoding from different sources should maintain semantic consistency; (b): Different views naturally have differences in describing the same label.}
\vspace{-0.2cm}
\label{fig:illu}
\end{figure}

From the perspective of multi-view representation learning, a fundamental issue is how to learn a common semantic representation under incomplete observations. Most existing iM3C methods are built around a representation-centric philosophy through shared representation learning, consistency regularization, information bottleneck objectives, or variational latent modeling \cite{wang2021learning,han2022trusted}: they first construct a common latent representation from incomplete views and then apply a multi-label classifier on top of it \cite{liu2024partial,wen2025learning}. This paradigm is effective for mitigating missing-view perturbations, since a shared representation can aggregate the target-relevant semantics observed from different views \cite{wang2026robust,liu2026learning}. However, such a design also introduces an implicit assumption: once a sufficiently good common representation is obtained, all labels can be predicted from the same fused evidence. This assumption is often too restrictive for multi-label scenarios.

In multi-label prediction, different semantic labels may rely on different evidence sources. For example, in the diagnosis of multiple eye diseases, views derived from RGB images are generally more suitable for detecting many typical visible lesions of diabetic retinopathy, whereas views describing Optical Coherence Tomography (OCT) images are more beneficial for identifying diabetic macular edema. Even for the same label, the most reliable view may vary from one instance to another due to noise, occlusion, or missing observations, which is commonly overlooked by existing multi-label fusion methods \cite{liu2024attention,black2024multi,gan2025multi}. Therefore, we argue that the relationship between views and labels is not globally fixed, but instance-dependent and label-specific. We refer to this phenomenon as view-label heterogeneity, as shown in Fig. \ref{fig:illu}(b).

These observations indicate that the problem should be addressed through two complementary mechanisms. The first one aims to learn a robust shared semantic space from incomplete observations, and the latter explicitly evaluates how much each available view contributes to each label prediction, so that label-specific complementary evidence is not suppressed by global fusion. Motivated by this, we develop a unified framework (V2L) that combines semantic consistency learning with view-label heterogeneity modeling. {To be specific, we propose a source-perturbation invariant encoding scheme to preserve task-relevant information and strengthen cross-view semantic consistency under incomplete observations, as shown in Fig. \ref{fig:illu}(a). It is worth noting that when performing perturbation alignment, we seek the greatest diversity}. On the decision side, we design an active view-label relevance evaluation strategy that evaluates instance-wise and label-aware view contributions for adaptive predictive fusion. In this way, mid-level semantic fusion and late-stage decision fusion are integrated rather than treated as isolated alternatives. 
The resulting framework establishes a coherent path from multi-view distribution consistency modeling to adaptive label-aware decision aggregation. It preserves the advantages of shared latent learning while avoiding the limitation that all labels must rely on a single fused representation or on static view weighting. Consequently, the proposed method provides a unified solution that simultaneously enhances semantic richness, cross-view consistency, and label-sensitive predictive adaptation for incomplete multi-view multi-label classification.

The main contributions of this work are summarized as follows:
\begin{itemize}
\item We revisit incomplete multi-view multi-label classification from a representation-decision perspective, and identify view-label heterogeneity as a key limitation of representation-only semantic fusion.
\item We present \textbf{V2L}, a hybrid framework for iM3C, which jointly models common semantic consistency and view-label heterogeneity, thereby bridging representation-level fusion and decision-level fusion in a unified paradigm.
\item We develop a source-perturbation invariant encoding mechanism to preserve task-relevant shared information and enforce robust cross-view semantic consistency under simultaneous view missingness and label incompleteness.
\item We propose an active view-label heterogeneity modeling strategy to characterize sample-specific and label-aware view contributions, enabling adaptive predictive fusion beyond static or globally learnable weighting schemes.
\end{itemize}

\section{Related Work}
\subsection{Multi-View Learning}
Multi-view learning aims to jointly exploit multiple representations of the same instance, where different views may arise from distinct feature extractors, modalities, or data sources \cite{wen2022survey,zhao2024active,zhou2023clustering}. By leveraging both shared and complementary information across views, it often yields more robust and discriminative representations than single-view learning. In unsupervised scenarios, multi-view clustering has become one of the most representative directions. Early and influential studies mainly focused on enforcing cross-view consistency through co-training or co-regularization \cite{kumar2011co}. {After that, multi-kernel and graph-based approaches have shown strong ability to integrate heterogeneous views \cite{yang2024auto,wang2019gmc}, while subspace-based methods are especially effective for high-dimensional data with latent low-rank structure \cite{chen2021low}. Recent graph-based developments have further explored high-order cross-view correlations through tensorized adaptive consensus graph learning \cite{guo2024tensor}, while anchor graph propagation has been employed to improve the scalability of clustering under incomplete multi-view observations \cite{pu2026conquering}. Recently, deep variational multi-view learning methods demonstrate outstanding performance while providing a solid theoretical foundation \cite{tian2024variational,khattar2019mvae,liu2024partial}.}

Another important direction concerns incomplete multi-view learning, where one or more views may be missing during training or inference \cite{ding2026incomplete,jiang2026reliable}. To alleviate the effect of missing observations. Existing methods mainly eliminate the impact of missing views through two approaches: view imputation or missing skipping \cite{wen2022survey}. For example, Liu et al. \cite{liu2024attention} proposed to fill missing views in the latent space based on the aggregation of nearest neighbor graphs across samples; Lin et al. \cite{lin2021completer} adopted contrastive learning to pursue consistent features and conducted cross-prediction to recover missing views. View imputation methods inevitably increase the computational complexity and also risk introducing redundant noise. Conversely, the other approach typically introduces prior missing information and masks the missing views during forward and backward propagation \cite{tan2018incomplete,xie2023exploring}. 

\subsection{Multi-Label Classification}
Multi-label classification aims to assign multiple semantic labels to each instance and has been widely studied in visual recognition, recommendation, medical diagnosis, and text understanding. Compared with single-label classification, its core challenge lies not only in extracting discriminative features, but also in characterizing complex dependency structures among labels \cite{li2022learning,liu2024attention}. Recent research has moved beyond early problem-transformation strategies \cite{cherman2011multi} and increasingly focuses on more realistic and robust settings, such as weakly supervised label discovery \cite{kim2022large}, single-positive annotation \cite{xu2022one}, and prediction reliability \cite{gong2025diversity}. For instance, Li et al. \cite{li2024open} proposed an open-world multi-label text classification framework under extremely weak supervision, where the label space is not fully given in advance and must be progressively discovered from raw documents. Recent studies have also paid increasing attention to the trustworthiness of multi-label predictors. Chen et al. \cite{chen2024dynamic} introduced the multi-label confidence calibration task and proposed a dynamic correlation learning method to improve the reliability of confidence estimates in multi-label recognition. 
Nevertheless, most multi-label classification methods are developed under single-view or fully observed feature settings. Even when multiple modalities or views are available, many methods still emphasize global feature fusion while paying limited attention to the fact that different labels may rely on different evidence sources \cite{black2024multi,zhao2022non}. This limitation becomes more pronounced in multi-view scenarios, where some labels are naturally better explained by specific views than others. Therefore, for the problem studied in this paper, it is insufficient to only model label correlation or to only strengthen the final classifier. What is further required is a label-aware mechanism that can evaluate which view contributes more to which label, and do so in an instance-dependent manner.

\subsection{Incomplete Multi-View Missing Multi-Label Classification}
Incomplete multi-view missing multi-label classification (iM3C) lies at the intersection of incomplete multi-view learning and incomplete multi-label learning. Its goal is to perform reliable multi-label prediction when both view missingness and label incompleteness are present simultaneously \cite{li2022concise,zhang2022effective,liu2025reliable}. Early work was only developed for a single missing setting, making it difficult to handle the dual missing of views and labels simultaneously, like GLOCAL \cite{zhu2017multi}, C2AE \cite{yeh2017learning}, and CDMM \cite{zhao2021consistency}. The first known work for iM3C is IMvML \cite{tan2018incomplete}, which adopts matrix factorization to align the multi-view data to the common latent space. Recently, more works shift toward deep representation learning due to its powerful feature extraction capabilities, such as DICNet \cite{liu2023dicnet} and AIMNet \cite{liu2024attention}. These studies have established the basic learning paradigm for iM3C and demonstrated that combining multi-view complementarity with weak-label supervision can yield substantial benefits. 

However, current iM3C methods still exhibit two limitations that are directly related to our motivation. First, many methods prioritize building a robust shared representation but provide limited treatment of how task-relevant information should be preserved in a semantically sufficient manner under incomplete observations. Second, existing methods ignore the dependency between views and multiple labels, simply regarding iM3C as a stacking task of representation learning and multi-label classification. This cuts off the intrinsic connection between multiple views and multiple labels. In other words, existing methods failed to fully clarify the relationship between multi-view semantic consistency learning and label discrimination.

\section{Preliminary}
\subsection{Problem Definition and Notations}
Let $\{\mathbf{x}^{(v)}\}_{v=1}^{m}$ be $m$ random variables describing the same object from different views, where $\mathbf{x}^{(v)}\in\mathbb{R}^{d_v}$, and let $\mathbf{y}\in\{0,1\}^{c}$ be the corresponding multi-label random variable. Denote by $\mathcal{M}=\{1,2,\ldots,m\}$ and $\mathcal{C}=\{1,2,\ldots,c\}$ the complete index sets of views and labels, respectively. A dataset with $n$ observations (samples) is written as $\mathcal{D}=\{(\{x_i^{(v)}\}_{v=1}^{m},y_i)\}_{i=1}^{n}$, where $x_i^{(v)}$ and $y_i$ are samplings of variables $\mathbf{x}^{(v)}$ and $\mathbf{y}$.

Under the incomplete setting, not all views and labels are observed for every sample. For the $i$-th sample, we denote by $\mathcal{V}_i\subseteq\mathcal{M}$ the set of available views (or $\mathcal{V}$ when no specific sample is involved) and by $\mathcal{G}_i\subseteq\mathcal{C}$ the set of observed labels. For the convenience of formal representation, we also define two index matrices $\mathbf{W}\in \mathbb{R}^{n\times m}$ and $\mathbf{R}\in \mathbb{R}^{n\times c}$ to indicate which view or label is available:
\begin{equation}
\mathbf{W}_{i,v}=
\begin{cases}
1, & \text{if } v\in\mathcal{V}_i\\
0, & \text{otherwise}
\end{cases}
,
\mathbf{R}_{i,j}=
\begin{cases}
1, & \text{if } j\in\mathcal{G}_i\\
0, & \text{otherwise}
\end{cases}.
\end{equation}
The two forms are equivalent in the sense that
$\mathcal{V}_i=\{v\in\mathcal{M}\mid \mathbf{W}_{i,v}=1\},
\mathcal{G}_i=\{j\in\mathcal{C}\mid \mathbf{R}_{i,j}=1\}.$ Therefore, the goal of incomplete multi-view multi-label classification is to learn a predictor that maps incomplete multi-view observations to the complete label space,
\begin{equation}
f:\{x_i^{(v)}\}_{v\in\mathcal{V}_i}\mapsto y_i\in[0,1]^c,
\end{equation}
so that the semantic label can be accurately inferred under simultaneous view and label incompleteness.

\subsection{Task-Relevant Common Semantics in Multi-View Data}

Although different views are obtained from heterogeneous sources or feature extractors, they describe the same underlying instance. Therefore, multi-view observations are expected to share certain semantic information that reflects the common properties of the target. Such common semantic information provides the basic foundation for learning a unified multi-view representation under incomplete observations.

To formalize this idea, we introduce a conceptual latent variable $\mathbf{s}$ to denote the common semantic information shared by multiple views. For each view $\mathbf{x}^{(v)}$, $\mathbf{s}$ represents the semantic factors that can be supported by the corresponding observation. This leads to the following assumption.

\begin{assumption}[Task-related common semantics]
\label{assumption}
For multi-view observations describing the same instance, there exists a latent semantic variable $\mathbf{s}$ that is shared by the views and related to the label variable $\mathbf{y}$. Formally, for each view $v\in\mathcal{M}$,
\begin{equation}
I(\mathbf{x}^{(v)};\mathbf{s})>0,\qquad I(\mathbf{s};\mathbf{y})>0.
\end{equation}
\end{assumption}

Here, $\mathbf{s}$ is used only as a conceptual variable to describe the common semantic factor in multi-view data, rather than an additional latent variable explicitly implemented in the model. Under this assumption, a desirable multi-view representation $\mathbf{z}$ should preserve the information carried by $\mathbf{s}$. This motivates the following proposition.

\begin{proposition}
\label{proposition1}
If a multi-view representation $\mathbf{z}$ preserves the common semantic information $\mathbf{s}$ shared by multiple views, then $\mathbf{z}$ can serve as a common semantic representation of the observed views.
\end{proposition}

\subsection{Multi-view Information Bottleneck Framework}
In principle, a properly constructed cross-view representation $\mathbf{z}$ can capture the information necessary for downstream prediction. However, raw multi-view observations often contain substantial view-specific variations that are irrelevant to the target. As a result, naive fusion may introduce such irrelevant factors into $\mathbf{z}$, leading to redundant information that obscures the shared task-relevant signal and degrades the predictability of $\mathbf{z}$ for $\mathbf{y}$ \cite{federici2020learning,liu2024partial}. These observations motivate learning strategies that emphasize shared semantics while suppressing view-specific irrelevant information, thereby better preserving the sufficiency of $\mathbf{z}$ under realistic conditions.

According to the Data Processing Inequality, the Markov chain
$\{\mathbf{x}^{(v)}\}_{v\in\mathcal{V}} \rightarrow \mathbf{z} \rightarrow \mathbf{y}$
implies the mutual information relationship among $\{\mathbf{x}^{(v)}\}_{v\in\mathcal{V}}, \mathbf{z}$, and $\mathbf{y}$:
\begin{equation}
I(\mathbf{z};\mathbf{y}) \leq I(\{\mathbf{x}^{(v)}\}_{v\in\mathcal{V}};\mathbf{y}).
\end{equation}
Hence, if the encoding process does not lose any information useful for predicting $\mathbf{y}$, then $\mathbf{z}$ is sufficient for $\mathbf{y}$ \cite{achille2018emergence}, namely,
\begin{equation}
I(\mathbf{z};\mathbf{y}) = I(\{\mathbf{x}^{(v)}\}_{v\in\mathcal{V}};\mathbf{y}).
\end{equation}

{However, sufficiency alone is not enough, since a redundant representation may still preserve large amounts of view-specific nuisance information. Following the information bottleneck principle, the desired representation should satisfy both sufficiency and compactness, which leads to
\begin{equation}
\label{eq:pre_ib}
\min I(\{\mathbf{x}^{(v)}\}_{v\in\mathcal{V}};\mathbf{z}) - I(\mathbf{z};\mathbf{y}).
\end{equation}
Eq. (\ref{eq:pre_ib}) is the natural form of the information bottleneck theory under the condition of multi-view learning. The first term can be understood as the compression process of information encoding, while the second term represents strengthening the correlation between latent encoding and the task. }

{However, it should be noted that the above sufficiency assumptions are not invulnerable. Although we expect different views to semantically describe the target consistently, from an information perspective, models built on such ideal assumptions may lead to the neglect of view-specific information. Therefore, in this paper, in addition to pursuing semantic consistency among multiple views by applying the multi-view information bottleneck theory, we also propose to model the view-label heterogeneity to supplement view-specific cues.}

\section{Methodology}
\begin{figure*}[h!]
\centering
\includegraphics[width=0.97\textwidth]{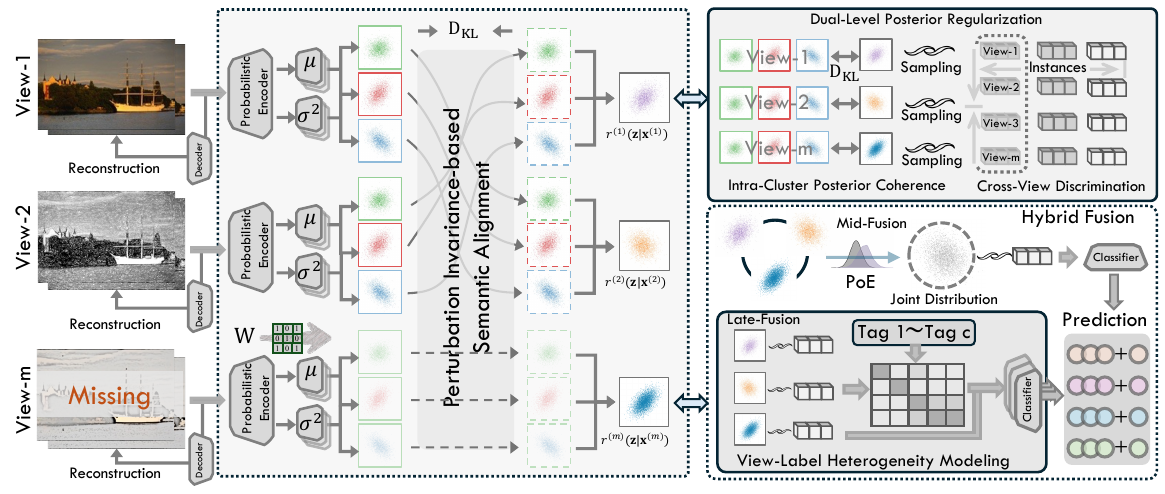} 
\caption{The main architecture diagram of our V2L. We adopt a multi-view variational encoder as the main framework. Cross-view semantic consistency is achieved by maintaining the distribution unchanged by exchanging the corresponding elements among distribution clusters. The decisions on each label will rely on different views, jointly forming a multi-view hybrid fusion structure. The missing index matrices are introduced to prevent unavailable views and labels from participating in the training.}
\label{fig:main}
\vspace{-0.3cm}
\end{figure*}

In this section, we first present our overall information theory-based objective, and then propose four components for optimizing this objective: source-perturbation invariant encoding, dual-level posterior regularization, active view-label relevance modeling, and multi-label classification. The overall framework is shown in Fig. \ref{fig:main}.
\subsection{Overall Objective}
We begin from the observation that incomplete multi-view learning is essentially a problem of shared semantic extraction under missing observations. Let $\mathbf{x}^{(u)}$ and $\mathbf{x}^{(v)}$ denote two observed views with $u\neq v$, and let $\mathbf{z}$ denote the joint multi-view representation. To characterize when $\mathbf{z}$ is sufficient for downstream prediction, we consider the mutual-information decomposition:
\begin{equation}
\begin{aligned}
I(\mathbf{x}^{(u)}; \mathbf{x}^{(v)}, \mathbf{z}) =&\; I(\mathbf{x}^{(u)};\mathbf{z}\mid\mathbf{x}^{(v)}) + I(\mathbf{x}^{(v)};\mathbf{x}^{(u)}) \\
=&\; I(\mathbf{x}^{(v)};\mathbf{x}^{(u)}\mid\mathbf{z}) + I(\mathbf{z};\mathbf{x}^{(u)}),
\end{aligned}
\end{equation}
which leads to
\begin{equation}\small
\label{eq:mi_decomp}
I(\mathbf{x}^{(u)};\mathbf{z}\mid\mathbf{x}^{(v)})
=
\underbrace{I(\mathbf{z};\mathbf{x}^{(u)})-I(\mathbf{x}^{(v)};\mathbf{x}^{(u)})}_{\text{minimality gap}}
+
\underbrace{I(\mathbf{x}^{(v)};\mathbf{x}^{(u)}\mid\mathbf{z})}_{\text{consistency gap}}.
\end{equation}

Eq.~(\ref{eq:mi_decomp}) reveals two desirable conditions for multi-view representation learning. First, if $\mathbf{z}$ contains all the information shared across views, then conditioning on $\mathbf{z}$ should remove the residual dependence between views, i.e., $I(\mathbf{x}^{(v)};\mathbf{x}^{(u)}\mid\mathbf{z})=0$. Second, if $\mathbf{z}$ contains only the shared information, then its mutual information with each view should match the shared part between views, i.e., $I(\mathbf{z};\mathbf{x}^{(u)})-I(\mathbf{x}^{(v)};\mathbf{x}^{(u)})=0$. Under these two conditions, the conditional mutual information $I(\mathbf{x}^{(u)};\mathbf{z}\mid\mathbf{x}^{(v)})$ vanishes, which formalizes the notion of minimal sufficiency with respect to cross-view common semantics. 
\begin{tcolorbox}[colback=gray!10,
                  colframe=black,
                  width=\linewidth,
                  arc=1mm, auto outer arc,
                  boxrule=0.5pt,
                 ]
\begin{corollary}
\label{cor:minsuff}
For any pair of views $u\neq v$, suppose that $\mathbf{z}$ is a minimal representation that preserves the task-related common semantic information shared by $\mathbf{x}^{(u)}$ and $\mathbf{x}^{(v)}$. Then conditioning on $\mathbf{x}^{(v)}$ makes $\mathbf{x}^{(u)}$ provide no extra information about $\mathbf{z}$, i.e.,
\[
I(\mathbf{x}^{(u)};\mathbf{z}\mid\mathbf{x}^{(v)})=0.
\]
\end{corollary}
\end{tcolorbox}
This corollary provides the principle behind our representation learning objective: a desirable latent representation should preserve valid information from each view while suppressing view-private redundancy that is not supported by other views. 
Therefore, combining Corollary \ref{cor:minsuff} and the information bottleneck design principle, we have the following representation learning objective:
\begin{equation}\small
\max\;
\frac{1}{|\mathcal{V}|}\sum_{v\in\mathcal{V}} I\big(\mathbf{z};\mathbf{x}^{(v)}\big),\;
\text{s.t.}\;
\frac{1}{|\mathcal{V}|}\sum_{\substack{u,v\in\mathcal{V}}}^{u\neq v}
I\big(\mathbf{x}^{(u)};\mathbf{z}\mid\mathbf{x}^{(v)}\big)=0.
\end{equation}
Introducing a Lagrange multiplier $\beta\ge 0$, we obtain:
\begin{equation}\small
\label{eq:rep_obj}
\max\;
\frac{1}{|\mathcal{V}|}\sum_{v\in\mathcal{V}} I\big(\mathbf{z};\mathbf{x}^{(v)}\big)
-
\beta\cdot
\frac{1}{|\mathcal{V}|}\sum_{\substack{u,v\in\mathcal{V}}}^{u\neq v}
I\big(\mathbf{x}^{(u)};\mathbf{z}\mid\mathbf{x}^{(v)}\big).
\end{equation}

Eq.~(\ref{eq:rep_obj}) captures the representation side of our framework, namely, preserving view-valid information while enforcing cross-view semantic consistency. Nevertheless, for incomplete multi-view multi-label classification, this objective is still incomplete because it does not explicitly account for task-relevant label information. To make the learned representation predictive for multi-label classification, we further maximize the mutual information between the common latent representation $\mathbf{z}$ and the target labels $\mathbf{y}$, which yields the overall objective:
\begin{equation}\small
\label{eq:overall_obj}
\begin{aligned}
\max\;
&\frac{1}{|\mathcal{V}|}\sum_{v\in\mathcal{V}} I(\mathbf{z};\mathbf{x}^{(v)})
-
\beta\cdot
\frac{1}{|\mathcal{V}|}\sum_{\substack{u,v\in\mathcal{V}}}^{u\neq v}
I(\mathbf{x}^{(u)};\mathbf{z}\mid\mathbf{x}^{(v)})\\
&+\lambda I(\mathbf{z};\mathbf{y}),
\end{aligned}
\end{equation}
where $\lambda\ge 0$ balances representation consistency and task relevance.


\subsection{Source-Perturbation Invariant Cross-View Encoding}

Having formulated the overall objective, we next consider how to obtain a tractable variational form for its representation part. 
The desired latent representation should satisfy two requirements simultaneously. 
First, each latent component is expected to retain the information carried by its associated view (first term in Eq. (\ref{eq:overall_obj})). 
Second, whenever different views describe the same underlying semantics, the corresponding latent posteriors should remain consistent across views (second term in Eq. (\ref{eq:overall_obj})). 
These two requirements naturally lead to a variational encoding scheme with cross-view posterior regularization.

Unlike existing variational frameworks that simply use single probabilistic encoder (e.g., multi-layer perceptron (MLP)) to directly model a specific distribution to approximate $p(\mathbf{z}|\mathbf{x}^{(v)}) \approx r^{(v)}(\mathbf{z}|\mathbf{x}^{(v)})$, we employ a cluster of probabilistic encoders to map the source data of each view to the latent space of all views. For each available source view $\mathbf{x}^{(v)}$, a group of cross-view encoders is employed to model a cluster of source-anchored distribution proposals:
\begin{equation}
\{r^{v\rightarrow l}(\mathbf{z} \mid \mathbf{x}^{(v)})\}_{l\in\mathcal{M}},
\label{eq:src_tgt_posterior}
\end{equation}
where $r^{v\rightarrow l}(\mathbf{z} | \mathbf{x}^{(v)})$ denotes a Gaussian posterior proposal over the shared latent variable $\mathbf{z}$, produced by the $l$-th target-view-aware encoder head from the $v$-th source view. All proposal posteriors are defined on the same shared latent
space. The index \(l\) specifies the target-view-aware proposal
head, rather than a separate latent variable.
These proposals provide multiple latent projections induced by the same observed view, so that the information carried by $\mathbf{x}^{(v)}$ is examined from different cross-view perspectives rather than from a single encoder path alone.

Given a cluster derived from $\mathbf{x}^{(v)}$, we further consolidate this proposal cluster into a single view-specific posterior $r^{(v)}(\mathbf{z} | \mathbf{x}^{(v)})$, denoted by:
\begin{equation}\small
r^{(v)}(\mathbf{z} | \mathbf{x}^{(v)})
=\mathcal{A}(\{r^{v\rightarrow l}(\mathbf{z}|\mathbf{x}^{(v)})\}_{l\in \mathcal{M}}):=
\mathcal{N}(\boldsymbol{\mu}^{(v)},\boldsymbol{\Sigma}^{(v)}),
\label{eq:poe_fusion}
\end{equation}
where $\mathcal{A}(\cdot)$ denotes the precision-weighted Gaussian aggregation over the proposal cluster and the parameters $\boldsymbol{\mu}^{(v)}$ and $\boldsymbol{\Sigma}^{(v)}$ are obtained by:
\begin{equation}\small
\begin{aligned}
&\boldsymbol{\Sigma}^{(v)}
=
(
\mathbf{I}
+
\sum_{l\in\mathcal{M}}
(\boldsymbol{\Sigma}^{v\rightarrow l})^{-1}
)^{-1},\\
&\boldsymbol{\mu}^{(v)}
=
\boldsymbol{\Sigma}^{(v)}
(
\sum_{l\in\mathcal{M}}
(\boldsymbol{\Sigma}^{v\rightarrow l})^{-1}
\boldsymbol{\mu}^{v\rightarrow l}
),
\end{aligned}
\label{eq:pwga_fusion}
\end{equation}
where $r^{v\rightarrow l}(\mathbf{z}| \mathbf{x}^{(v)}):=\mathcal{N}(\boldsymbol{\mu}^{v\rightarrow l},\boldsymbol{\Sigma}^{v\rightarrow l})$ and $\mathbf{I}$ corresponds to the standard Gaussian prior. 
{Eq.~(\ref{eq:pwga_fusion}) assigns larger weights to proposals with smaller posterior variance, thereby suppressing uncertain projections and retaining more reliable semantic cues. 
In this way, $r^{(v)}(\mathbf{z}| \mathbf{x}^{(v)})$ serves as the consolidated posterior expert associated with view $v$. This ``divergence-aggregation'' flow ($v\rightarrow \mathcal{M} \rightarrow v$) provides the foundation for our perturbation-consistency alignment.}

\subsubsection{Variational Realization of Cross-View Semantic Consistency}

Returning to the first two terms in Eq.~(\ref{eq:overall_obj}), we first derive a tractable objective for preserving view-valid information. For each available view $v\in\mathcal{V}$, the term $I(\mathbf{z};\mathbf{x}^{(v)})$ measures how much information the latent posterior retains about its source observation. By the definition of mutual information, we have:
\begin{equation}\small
\begin{aligned}
&I(\mathbf{x}^{(v)};\mathbf{z})\\
&=\!
\iint \! p(\mathbf{x}^{(v)},\mathbf{z})
\log \frac{p(\mathbf{x}^{(v)}| \mathbf{z})}{p(\mathbf{x}^{(v)})}
\,d\mathbf{x}^{(v)}d\mathbf{z} \\
&=\!
\iint \!p(\mathbf{x}^{(v)},\mathbf{z})
\log p(\mathbf{x}^{(v)}| \mathbf{z})
\,d\mathbf{x}^{(v)}d\mathbf{z}
\!+\!
H(\mathbf{x}^{(v)}).
\end{aligned}
\end{equation}
Since $H(\mathbf{x}^{(v)})\geq 0$, maximizing $I(\mathbf{x}^{(v)};\mathbf{z})$ can be lower-bounded by maximizing the conditional log-likelihood of reconstructing $\mathbf{x}^{(v)}$ from $\mathbf{z}$. Introducing a variational decoder $q^{(v)}(\mathbf{x}^{(v)}\mid \mathbf{z})$, we obtain:
\begin{equation}\small
\label{eq:view_recon_lb}
I(\mathbf{x}^{(v)};\mathbf{z})
\ge
\mathbb{E}_{p(\mathbf{x}^{(v)})}
\mathbb{E}_{r^{(v)}(\mathbf{z}\mid \mathbf{x}^{(v)})}
\big[\log q^{(v)}(\mathbf{x}^{(v)}\mid \mathbf{z})\big].
\end{equation}
Eq.~(\ref{eq:view_recon_lb}) shows that a reconstruction objective naturally serves as a variational surrogate for the first term in Eq.~(\ref{eq:overall_obj}). Intuitively, this term prevents the latent variable from being overly compressed and preserves the information in each observed view.

For the second term in Eq.~(\ref{eq:overall_obj}), we consider the conditional mutual information between $\mathbf{x}^{(u)}$ and $\mathbf{z}$ given $\mathbf{x}^{(v)}$, where $u\neq v$:
\begin{equation}\footnotesize
\begin{aligned}
&I(\mathbf{x}^{(u)};\mathbf{z}\mid \mathbf{x}^{(v)})\\
&=
\iiint p(\mathbf{x}^{(u)},\mathbf{x}^{(v)},\mathbf{z})
\log
\frac{p(\mathbf{z}\mid \mathbf{x}^{(u)},\mathbf{x}^{(v)})}
{p(\mathbf{z}\mid \mathbf{x}^{(v)})}
\, d\mathbf{x}^{(u)} d\mathbf{x}^{(v)} d\mathbf{z}\\
&=\mathbb{E}_{p(\mathbf{x}^{(u)},\mathbf{x}^{(v)})}
\left[D_{\mathrm{KL}}
\left(
p(\mathbf{z}\mid \mathbf{x}^{(u)},\mathbf{x}^{(v)})
\Vert
p(\mathbf{z}\mid \mathbf{x}^{(v)})
\right)\right].
\end{aligned}
\end{equation}
{This expression indicates that reducing $I(\mathbf{x}^{(u)};\mathbf{z}\mid \mathbf{x}^{(v)})$ amounts to making the posterior of $\mathbf{z}$ conditioned on both views close to the posterior conditioned only on view $v$. Since the exact conditional mutual information involves the intractable posteriors $p(\mathbf{z}\mid\mathbf{x}^{(u)},\mathbf{x}^{(v)})$ and $p(\mathbf{z}\mid\mathbf{x}^{(v)})$, it cannot be directly evaluated. We therefore introduce the following conditional mutual information (CMI)-motivated posterior-consistency surrogate:
\begin{equation}\small
\label{eq:cmi_upper_pami}
\mathcal{S}_{\mathrm{CMI}}^{(u,v)}
=
\sum_{l\in\mathcal{M}}
D_{\mathrm{KL}}\!\left(
r^{u\rightarrow l}(\mathbf{z}\mid\mathbf{x}^{(u)})
\,\middle\|\,
r^{v\rightarrow l}(\mathbf{z}\mid\mathbf{x}^{(v)})
\right).
\end{equation}
The proposals compared in Eq.~(\ref{eq:cmi_upper_pami}) are generated from different source views but correspond to the same target-aware head $l$. Minimizing their discrepancy encourages the target-wise semantic posterior to remain invariant to the source view, which is consistent with the desired zero-CMI condition. We emphasize that $\mathcal{S}_{\mathrm{CMI}}^{(u,v)}$ is an optimizable consistency surrogate rather than an exact estimator or a variational bound of the conditional mutual information.}

{Based on Eq.~(\ref{eq:view_recon_lb}) and the CMI-motivated posterior-consistency surrogate in Eq.~(\ref{eq:cmi_upper_pami}), we instantiate the representation learning objective using the following information bottleneck loss $\mathcal{L}_{\mathrm{ib}}$:}
\begin{equation}\small
\label{eq:lib_pami}
\mathcal{L}_{\mathrm{ib}}
=
\mathcal{L}_{\mathrm{re}}
+
\beta\,\mathcal{L}_{\mathrm{consis}},
\end{equation}
where
\begin{equation}\small
\label{eq:lre_pami}
\mathcal{L}_{\mathrm{re}}
=
-\frac{1}{|\mathcal{V}|}
\sum_{v\in\mathcal{V}}
\mathbb{E}_{r^{(v)}(\mathbf{z}\mid \mathbf{x}^{(v)})}
\big[
\log q^{(v)}(\mathbf{x}^{(v)}\mid \mathbf{z})
\big],
\end{equation}
and
\begin{equation}\small
\label{eq:lpc_pami}
\mathcal{L}_{\mathrm{consis}}
=
\frac{1}{|\mathcal{V}|}
\sum_{\substack{u,v\in\mathcal{V}}}^{u\neq v}
\sum_{l\in\mathcal{M}}D_{\mathrm{KL}}\big(
r^{v\rightarrow l}(\mathbf{z}\mid \mathbf{x}^{(v)})
\,\|\, 
r^{u\rightarrow l}(\mathbf{z}\mid \mathbf{x}^{(u)})
\big).
\end{equation}
Here $\mathcal{L}_{\mathrm{re}}$ preserves view-valid information and prevents over-compression, while $\mathcal{L}_{\mathrm{consis}}$ encourages the latent posterior of each target view to be consistently induced from different observed views. Their combination provides a practical realization of minimum sufficiency under incomplete multi-view observations.

\subsubsection{Source-Perturbation Invariance Modeling}

Although the regularizer $\mathcal{L}_{\mathrm{consis}}$ in Eq.~(\ref{eq:lpc_pami}) directly encourages cross-view semantic consistency, a full enumeration of all ordered view pairs may introduce unnecessary computational burden and may also make the alignment process overly rigid. To obtain a more scalable and robust mechanism, we further introduce a perturbation invariance strategy based on random exchange over the source-anchored posterior clusters rather than iteration.

For a given available source view $v$, recall that it induces a semantic proposal cluster
$\{r^{v\rightarrow l}(\mathbf{z}\mid \mathbf{x}^{(v)})\}_{l\in\mathcal{M}}$.
We denote the latent proposal cluster $\mathcal{C}^{(v)}$ by:
\begin{equation}
\mathcal{C}^{(v)}=\{
r^{v\rightarrow l}(\mathbf{z}\mid \mathbf{x}^{(v)})
\}_{l\in\mathcal{M}}.
\end{equation}
Each cluster can be viewed as a structured semantic description induced from one observed view and projected onto all target latent subspaces. If different views encode the same underlying semantics, then the corresponding elements in these clusters should remain distributionally consistent even when the source-view association is perturbed.

Based on this intuition, for each sample, we randomly generate a perturbation mapping,
$\pi:\mathcal{V}\rightarrow\mathcal{V}$ over the set of available views, where $\pi(v)\neq v$.
We then replace the original source association of the $l$-th proposal by its counterpart and form a perturbed cluster:
\begin{equation}
\widetilde{\mathcal{C}}^{(v)}=\{
r^{\pi(v)\rightarrow l				}(\mathbf{z}\mid \mathbf{x}^{(\pi(v))})
\}_{l\in\mathcal{M}}.
\end{equation}

Under the minimum-sufficiency principle introduced above, if the semantic content encoded by different observed views is consistent at the target-relevant level, then permuting the source-view identity should not change the target-wise latent semantics. This observation leads to the following corollary.

\begin{tcolorbox}[colback=gray!10,
                  colframe=black,
                  width=\linewidth,
                  arc=1mm, auto outer arc,
                  boxrule=0.5pt]
\begin{corollary}[Source-perturbation invariance]
\label{cor:perm_consistency}
If two views are semantically equivalent with respect to the target-relevant information, then exchanging their roles as the source view does not alter the induced proposal cluster, i.e., $\mathcal{C}^{(v)}$ and $\widetilde{\mathcal{C}}^{(v)}$ are equivalent in target-wise distribution. Thus for $\forall\, l\in\mathcal{M}$, we have:
\[\small
D_{\mathrm{KL}}(
r^{v\rightarrow l}(\mathbf{z}\mid \mathbf{x}^{(v)})
\|
r^{\pi(v)\rightarrow l}(\mathbf{z}\mid \mathbf{x}^{(\pi(v))})
)=0.
\]
\end{corollary}
\end{tcolorbox}
{Moreover, the exchange scheme is designed to promote diversity across targets. For a fixed source view $v$, the exchanged sources assigned to different targets are required to be non-repeated:
\begin{equation}
\pi_l(v)\neq \pi_{l'}(v),\qquad \forall\, l\neq l',\quad l,l'\in\mathcal{M},
\end{equation}
where $\pi_l(v)$ and $\pi_l'(v)$ denote $l$- and $l'$-th random mapping.
Equivalently,
$
\bigl|\{\pi_l(v): l\in\mathcal{M}\}\bigr| = |\mathcal{M}|.
$
Therefore, the target-wise proposals are matched with semantically consistent yet diverse source anchors, which avoids repeatedly contrasting the same exchanged view across different targets and enriches the supervision for cluster-level semantic alignment.}
To sum up, instead of exhaustively matching all ordered view pairs, we rewrite $\mathcal{L}_{\mathrm{consis}}$ to regularize the discrepancy between the original cluster and its permuted counterpart: 
\begin{equation}\small
\label{eq:perm_consistency}
\begin{aligned}
&\mathcal{L}_{\mathrm{consis}}\\
&=
\frac{1}{|\mathcal{V}|}
\sum_{v\in\mathcal{V}}
\sum_{l\in\mathcal{M}}
D_{\mathrm{KL}}\!\big(
r^{v\rightarrow l}(\mathbf{z}| \mathbf{x}^{(v)})
\|
r^{\pi(v)\rightarrow l}(\mathbf{z}| \mathbf{x}^{(\pi(v))})
\big).
\end{aligned}
\end{equation}

It is worth emphasizing that the proposed perturbation-aware consistency loss does not assume that all views are identical at the raw-input level. Instead, it only requires their latent semantic proposals to remain consistent at the target-relevant level. Therefore, the model is still allowed to preserve moderate view-specific information through the reconstruction pathway, while the perturbation-aware consistency regularization suppresses view-private variations. This property is particularly important under incomplete observations, where directly enforcing hard alignment may amplify noise from low-quality or partially missing views. By contrast, our perturbation consistency modeling provides a softer and more efficient way to promote semantic cohesion across views while maintaining sufficient flexibility for practical multi-view multi-label prediction.

\subsection{Dual-Level Posterior Regularization}

While the source-perturbation invariant encoding promotes distributional consistency among source-anchored semantic proposals, it does not explicitly constrain the posterior coherence within each proposal cluster or the instance-level discriminability among different observed views. To further stabilize latent semantic learning, we introduce a dual-level posterior regularization scheme, which consists of intra-cluster posterior coherence regularization and cross-view instance-level posterior discrimination.

\paragraph{Intra-cluster posterior coherence.}
For a given observed view $v\in\mathcal{V}$, the source-anchored proposal cluster $\{r^{v\rightarrow l}(\mathbf{z}| \mathbf{x}^{(v)})\}_{l\in\mathcal{M}}$ contains multiple latent proposals induced from the same source observation. To reduce view-specific drift within the proposal cluster, we encourage each source-induced proposal to align with the corresponding view-specific posterior $r^{(v)}(\mathbf{z}\mid \mathbf{x}^{(v)})$:
\begin{equation}\small
\label{eq:lintra_pami}
\mathcal{L}_{\mathrm{intra}}
=
\frac{1}{|\mathcal{V}|}
\sum_{v\in\mathcal{V}}
\sum_{l\in\mathcal{M}}
D_{\mathrm{KL}}\!\big(
r^{v\rightarrow l}(\mathbf{z}\mid \mathbf{x}^{(v)})
\,\|\, 
r^{(v)}(\mathbf{z}\mid \mathbf{x}^{(v)})
\big).
\end{equation}

\paragraph{Cross-view posterior discrimination.}
Beyond intra-cluster posterior coherence, different observed views of the same sample should remain close after semantic encoding, while representations from different samples should stay distinguishable. To this end, we regularize the view-level posterior means through an instance-aware cross-view discrimination objective.
Let $\boldsymbol{\mu}_{i}^{(v)}\in\mathbb{R}^{d}$ be the posterior mean associated with the $i$-th sample under view $v$, and let $\tilde{\boldsymbol{\mu}}_{i}^{(v)}=f_{l_2}(\boldsymbol{\mu}_{i}^{(v)})$ be its $l_2$ normalized form. For two samples $i$ and $j$ under views $v$ and $u$, respectively, we define
$
s_{i,j}^{(v,u)}
=
\frac{
\langle
\tilde{\boldsymbol{\mu}}_{i}^{(v)},
\tilde{\boldsymbol{\mu}}_{j}^{(u)}
\rangle
}{\tau},
$
where $\tau>0$ is a temperature parameter.
Then the cross-view posterior discrimination loss is written as:
\begin{equation}
\label{eq:linter_new}
\mathcal{L}_{\mathrm{cross}}
=
-\frac{1}{n}
\sum_{i=1}^{n}
\sum_{v,u=1}^{m}
\mathbf{W}_{i,v}\mathbf{W}_{i,u}
\log
\frac{
\exp\!\big(s_{i,i}^{(v,u)}\big)
}{
\sum\limits_{j=1}^{n}
\exp\!\big(s_{i,j}^{(v,u)}\big)
\mathbf{W}_{j,u}
}.
\end{equation}
Eq. (\ref{eq:linter_new}) implements a contrastive posterior
regularizer: posteriors from different views of the same instance
are treated as positives, whereas posteriors from different
instances serve as negatives. This encourages cross-view semantic
consistency while preserving instance-level discriminability.

Overall, our posterior regularization loss $\mathcal{L}_{\mathrm{post}} = \sigma\mathcal{L}_{\mathrm{intra}} + \gamma\mathcal{L}_{\mathrm{cross}}$ complements the perturbation invariance regularization from two additional aspects: $\mathcal{L}_{\mathrm{intra}}$ improves the coherence of target-specific semantic proposals generated from the same source view, while $\mathcal{L}_{\mathrm{cross}}$ strengthens instance-level discriminability across different observed views. $\sigma$ and $\gamma$ are the trade-off coefficients.

\begin{figure}[t!]
\centering
\includegraphics[width=0.99\linewidth]{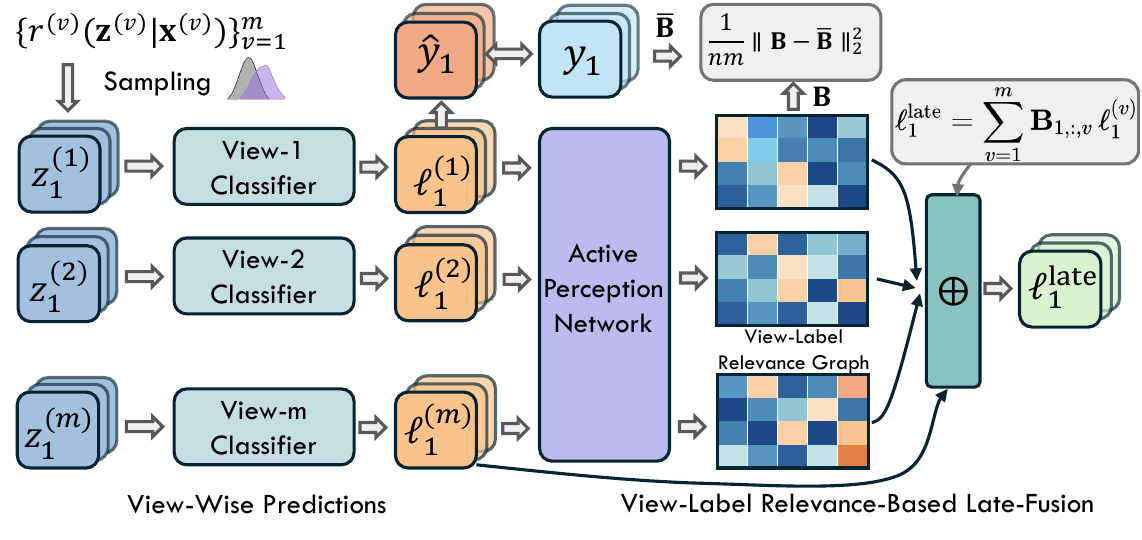} 
\caption{Schematic diagram of view-label heterogeneity modeling. The active perception network provides an instance-by-instance and view-by-view label-aware relevance score for the logits of each view.}
\label{fig:lqd}
\end{figure}
\subsection{Active View-Label Relevance Modeling}

The representation learning objectives above mainly target shared semantic sufficiency. However, returning to the overall objective Eq.~(\ref{eq:overall_obj}), the last term requires the learned representation to remain maximally informative about the target labels. A common realization is to maximize a variational lower bound of $I(\mathbf{z};\mathbf{y})$, which in practice reduces to minimizing the binary cross-entropy (BCE) between the ground-truth label and the model prediction. However, if this term is instantiated only through a single predictor built on the fused representation, then the optimization mainly rewards globally shared semantics and may overlook two important issues: (1) \textit{How different views contribute to final prediction?} (2) \textit{How to differentiate different labels?} The former reflects the need to explore the complementarity of multiple views, while the latter emphasizes the importance of transitioning from upstream representation learning to downstream multi-label classification. 

To cope with the two issues, in this section, we innovatively propose the view-label heterogeneity modeling strategy, which is motivated by the following observation: Different views contribute differently to different labels at the instance level. 
\begin{tcolorbox}[colback=gray!10,
                  colframe=black,
                  width=\linewidth,
                  arc=1mm, auto outer arc,
                  boxrule=0.5pt,
                 ]
\begin{definition}[View-label heterogeneity]  
For an instance \(i\), let \(s_{i,j}^{(v)}\) denote the predictive utility of view \(v\) for label \(j\). If there exist labels \(j_1 \neq j_2\) or instances \(i_1 \neq i_2\) such that
\[
s_{i,j_1}^{(v)} \neq s_{i,j_2}^{(v)}
\quad \text{or} \quad
s_{i_1,j}^{(v)} \neq s_{i_2,j}^{(v)},
\]
then the multi-view multi-label data exhibit view-label heterogeneity.
\end{definition}
\end{tcolorbox}
This definition implies that the view relevant to one label may be uninformative for another label, and such relevance can further vary from one instance to another. This requires us to move beyond simply mapping the joint representation of multiple views to the label space, and instead model the relevance between specific views and specific labels. To model such view-label heterogeneity with fine granularity, it is clearly insufficient to rely solely on mid-level multi-view fusion. Therefore, we introduce decision-level fusion while mapping joint features from multiple views to label prediction, forming a hybrid multi-view fusion framework that combines mid-level representation fusion and late-decision fusion.

For mid-level representation fusion, given view-specific approximate distribution $r^{(v)}(\mathbf{z} | \mathbf{x}^{(v)})$, we then construct the joint approximate posterior by applying a PoE operation to all available view-specific posteriors: 
\begin{equation}\label{eq:pzx_poe}
r(\mathbf{z}|\mathbf{x}^{\mathcal{V}})  := p(\mathbf{z})\prod_{v\in\mathcal{V}}r^{(v)}(\mathbf{z}|\mathbf{x}^{(v)}),
\end{equation}
where $\mathbf{x}^{\mathcal{V}} = \{\mathbf{x}^{(v)}\}_{v\in\mathcal{V}}$ and $p(\mathbf{z})$ is treated as the Standard Gaussian for convenience.
{In implementation, the joint latent representation used by the
mid-level fusion branch is sampled as
$z_i\sim r(\mathbf{z}|\mathbf{x}^{\mathcal{V}})$
whereas the latent representation associated with an available
view $v\in\mathcal V$ is sampled as $z_i^{(v)}\sim r^{(v)}(\mathbf{z}|\mathbf{x}^{(v)})$. Both $z_i$ and $z_i^{(v)}$ are obtained using the
reparameterization trick.} The joint representation is commonly used to produce a mid-level semantic prediction $\hat{y}_i^{\mathrm{mid}}$:
\begin{equation}
\ell^{\mathrm{mid}}_i
=
g(z_i),
\qquad
\hat{y}_i^{\mathrm{mid}}
=
f_{sig}(\ell^{\mathrm{mid}}_i),
\label{eq:mid_pred_pami}
\end{equation}
where $g(\cdot)$ denotes the shared classifier, $\ell_i^{\mathrm{mid}}$ is the logit score of mid-fusion, and $f_{sig}$ means the Sigmoid activation function. This branch mainly conveys the shared semantic evidence retained by source-perturbation invariant encoding.
Meanwhile, for late-level decision fusion, each view-specific representation is mapped to an individual decision branch:
\begin{equation}
\ell^{(v)}_i
=
g^{(v)}(z^{(v)}_i),
\qquad
\hat{y}^{(v)}_i
=
f_{sig}(\ell^{(v)}_i),
\quad v\in\mathcal{M},
\label{eq:view_pred_pami}
\end{equation}
where $\ell^{(v)}_i$, $g^{(v)}$, and $\hat{y}^{(v)}_i$ denote the logit, classifier, and prediction corresponding to view $v$, respectively.

Given these view-specific logits $\{\ell^{(v)}_i\}_{v\in \mathcal{M}}$, existing fusion strategies are clearly insufficient, since the contribution of a view is both label-dependent and sample-dependent. We therefore stack the view- and instance-wise logits as a tensor
$\mathbf{L}=\big[\ell^{(1)}_i,\dots,\ell^{(m)}_i\big]_{i=1}^n$, $\mathbf{L}\in\mathbb{R}^{n\times c\times m}.$
For the $j$-th label, the corresponding logit $\mathbf{L}_{i,j,:}$ is fed into an active perception network $f_{act}(\cdot)$ to estimate the relevance of each view to that label. The resulting relevance graph $\mathbf{B}\in \mathbb{R}^{n\times c\times m}$ is given by:
\begin{equation}\small
\mathbf{B}_{i,j,v}
=
\frac{
\exp\big(f_{act}(\mathbf{L}_{i,j,v})\big)\mathbf{W}_{i,v}
}{
\sum_{u=1}^{m}
\exp\big(f_{act}(\mathbf{L}_{i,j,u})\big)\mathbf{W}_{i,u}
},
\label{eq:perception_weight_pami}
\end{equation}
which excludes unavailable views from the normalization. It should be emphasized that our view-label heterogeneity aware approach is very different from both static late-fusion strategy and even view-level dynamic strategy that assigns equal weight to instances within the same view.

Based on the relevance graph, the late-fusion decision branch aggregates the view-wise logits in a label-aware manner:
\begin{equation}
\ell_{i}^{\mathrm{late}}
=
\sum_{v=1}^{m}
\mathbf{B}_{i,:,v}\,\ell_{i}^{(v)}.
\label{eq:late_pred_pami}
\end{equation}
The final prediction is then obtained by combining the mid-level semantic logits and the late-stage decision logits in the same logit space:
\begin{equation}
\hat{y}_i
=
f_{sig}\big(
\ell^{\mathrm{mid}}_i
+
\ell^{\mathrm{late}}_i
\big).
\label{eq:final_pred_pami}
\end{equation}
In this way, the mid-level branch supplies semantically stable shared evidence, whereas the late-stage branch preserves complementary label-specific contributions from individual views.

Here comes the next question: \textit{How to ensure that the output of the active perception network $f_{act}(\cdot)$ is effective}? Intuitively, the predictive effectiveness of a view for a given label should be directly reflected in its predictive performance. Therefore, we attempt to establish a self-guided supervision mechanism, that is, to supervise the learning of the active perception network via quality discrimination regarding the distribution modeled by the variational framework itself. 
Let $\mathbf{E}$ be prediction error tensor of views, whose element is calculated as:
\begin{equation}
\mathbf{E}_{i,j,v}
=
-\Big[
y_{i,j}\log \hat{y}_{i,j}^{(v)}
+
(1-y_{i,j})\log(1-\hat{y}_{i,j}^{(v)})
\Big].
\label{eq:view_bce_score_pami}
\end{equation}
We convert these errors into target relevance weights by:
\begin{equation}\small
\mathbf{\bar{B}}_{i,j,v}
=
\frac{
\exp(-\mathbf{E}_{i,j,v})\,\mathbf{W}_{i,v}
}{
\sum_{u=1}^{m}
\exp(-\mathbf{E}_{i,j,u})\,\mathbf{W}_{i,u}
},
\label{eq:target_weight_pami}
\end{equation}
so that a view with a smaller prediction error receives a larger target relevance. The resulting relevance regularization is
\begin{equation}\small
\mathcal{L}_{\mathrm{rele}}
=\frac{1}{n}\sum_{i=1}^{n}\frac{1}{|\mathcal{G}_i||\mathcal{V}_i|}\sum_{j=1}^{c}\mathbf{R}_{i,j}\sum_{v=1}^{m}\mathbf{W}_{i,v}(\mathbf{B}_{i,j,v}-\mathbf{\bar{B}}_{i,j,v})^{2}.
\label{eq:lrel_pami}
\end{equation}
{It is worth noting that, in order not to interfere with the main variational framework, the calculation of error $\mathbf{E}$ and target $\mathbf{\bar{B}}$ is only employed to supervise the learning of the active perceptron, thus ensuring the stability of training and the independence of inference.}

\begin{algorithm}[t]
\caption{Training Procedure of V2L}
\label{alg:v2l}
\begin{algorithmic}[1]
\State \textbf{Input:} Incomplete multi-view data $\mathcal{D}=\{(\{\mathbf{x}^{(v)}\}_{v=1}^{m},\mathbf{y})\}$, view index matrix $\mathbf{W}$, label index matrix $\mathbf{R}$, training epochs $T$, learning rate, and trade-off coefficients $\beta$, $\sigma$, and $\gamma$.
\State \textbf{Output:} The parameters of encoders, decoders, classifiers, and active perception network.
\For{$t=1,\ldots,T$}
		\State Compute source-anchored proposal cluster $\{r^{v\rightarrow u}(\mathbf{z}\mid \mathbf{x}^{(v)})\}_{u\in\mathcal{M}}$ for each available view.
		\State Aggregate proposal clusters by Eq.~(\ref{eq:poe_fusion}) to obtain view-specific posteriors $\{r^{(v)}(\mathbf{z}\mid \mathbf{x}^{(v)})\}_{v\in\mathcal{V}}$.
		\State {Construct the common posterior $r(\mathbf{z}\mid\mathbf{x}^{\mathcal{V}})$ by Eq.~(\ref{eq:pzx_poe}) and sample latent variables $\mathbf{z}$.}
		\State Reconstruct each observed view by decoders $q^{(v)}(\mathbf{x}^{(v)}\mid\mathbf{z})$ and compute $\mathcal{L}_{\mathrm{re}}$.
		\State Perform source-perturbation invariance matching and compute $\mathcal{L}_{\mathrm{consis}}$, $\mathcal{L}_{\mathrm{intra}}$, and $\mathcal{L}_{\mathrm{cross}}$.
		\State {Sample view-specific representation $z_i^{(v)}$ and joint representation $z_i$ from $r^{(v)}(\mathbf{z}\mid\mathbf{x}^{\mathcal{V}})$ and $r(\mathbf{z}\mid\mathbf{x}^{\mathcal{V}})$, respectively}.
		\State {Obtain the mid-level prediction $\hat{{y}}^{\mathrm{mid}}_i$ by Eq.~(\ref{eq:mid_pred_pami}) and the view-specific predictions $\{\hat{{y}}_i^{(v)}\}_{v=1}^{m}$ by Eq.~(\ref{eq:view_pred_pami}) for each sample $i$.}
		\State Build the logit tensor $\mathbf{L}$, infer the view-label relevance graph $\mathbf{B}$ by Eq.~(\ref{eq:perception_weight_pami}), and obtain the late-fusion logits by Eq.~(\ref{eq:late_pred_pami}).
		\State Compute the final prediction $\hat{{y}}_i$ by Eq.~(\ref{eq:final_pred_pami}), the target relevance graph $\bar{\mathbf{B}}$ by Eq.~(\ref{eq:target_weight_pami}), and the relevance loss $\mathcal{L}_{\mathrm{rele}}$.
		\State Compute the information bottleneck loss $\mathcal{L}_{\mathrm{ib}}$ and the posterior regularization loss $\mathcal{L}_{\mathrm{post}}=\sigma\mathcal{L}_{\mathrm{intra}}+\gamma\mathcal{L}_{\mathrm{cross}}$.
		\State Compute the classification loss $\mathcal{L}_{\mathrm{cls}}$ on observed labels and total objective $\mathcal{L}$ by Eq.~(\ref{eq:overall_train_obj_pami}).
		\State Update all network parameters by back-propagation with respect to $\mathcal{L}$.
\EndFor
\State \textbf{Return} the trained V2L model for inference.
\end{algorithmic}
\end{algorithm}
\vspace{-0.2cm}
\subsection{Multi-Label Classification and Overall Losses}

According to the label index matrix $\mathbf{R}$ defined in the problem formulation, only the labels in $\mathcal{G}_i$ are involved in supervised learning for the $i$-th sample. Therefore, the final multi-label classification loss is defined on the observed labels and hybrid prediction $\hat{y}_i$. Besides, since the adaptive decision branch is built upon the view-wise predictions, each single-view classifier should also retain sufficient predictive ability. Introducing the view index matrix $\mathbf{W}$, our overall multi-label classification loss $\mathcal{L}_{\mathrm{cls}}$ is formulated as:
\begin{equation}\small
\begin{aligned}
\mathcal{L}_{\mathrm{cls}}
&=
-\frac{1}{n}
\sum_{i=1}^{n}\frac{1}{|\mathcal{G}_i|}\sum_{j=1}^{c}
\mathbf{R}_{i,j}\big[
(
y_{i,j}\log \hat{y}_{i,j}
+
(1-y_{i,j})\log(1-\hat{y}_{i,j})
)\\
&+\frac{1}{|\mathcal{V}_i|}\sum_{v=1}^{m}
\mathbf{W}_{i,v}
(
y_{i,j}\log \hat{y}_{i,j}^{(v)}
+
(1-y_{i,j})\log(1-\hat{y}_{i,j}^{(v)}))\big].
\end{aligned}
\label{eq:ltask_pami}
\end{equation}

Combining information bottleneck loss $\mathcal{L}_{\mathrm{ib}}$, posterior regularization loss $\mathcal{L}_{\mathrm{post}}$, view-label relevance loss $\mathcal{L}_{\mathrm{rele}}$, and the classification loss $\mathcal{L}_{\mathrm{cls}}$, our overall training objective is given by:
\begin{equation}\small
\mathcal{L} 
=
\mathcal{L}_{\mathrm{cls}}
+
\mathcal{L}_{\mathrm{rele}}
+
\mathcal{L}_{\mathrm{ib}} 
+
\mathcal{L}_{\mathrm{post}}.
\label{eq:overall_train_obj_pami}
\end{equation}
where $\mathcal{L}_{\mathrm{ib}} = \mathcal{L}_{\mathrm{re}}+\beta\,\mathcal{L}_{\mathrm{consis}}$ and $\mathcal{L}_{\mathrm{post}} = \sigma\mathcal{L}_{\mathrm{intra}} + \gamma\mathcal{L}_{\mathrm{cross}}$ imply three trade-off coefficients. In Eq.~(\ref{eq:overall_train_obj_pami}), the first two terms instantiate the predictive realization of $I(\mathbf{z};\mathbf{y})$, while the remaining terms promote cross-view semantic consistency. In this way, the predictive objective and the representation objective are optimized within a unified framework.

To summarize the above components more clearly, the overall training procedure of V2L is given in Algorithm~\ref{alg:v2l}. The whole framework can be understood as two tightly coupled stages within each iteration: first, the model learns semantically sufficient and consistent latent representations from incomplete multi-view observations; second, it performs label-aware decision fusion based on the learned view-label relevance graph and updates all parameters by minimizing the unified objective.

{\subsection{Complexity analysis.}
Let $b$ denote the mini-batch size, and let $C_{\mathrm{enc}}$ and $C_{\mathrm{dec}}$ denote the costs of one probabilistic encoder and decoder, respectively. Since each available
source view is processed by $m$ target-aware encoder heads, proposal generation requires $\mathcal O(b|\mathcal V|mC_{\mathrm{enc}})$ operations. The reconstruction and posterior alignment terms require $\mathcal O(b|\mathcal V|C_{\mathrm{dec}})$ and $\mathcal O(b|\mathcal V|md)$ operations, respectively, while the batch-wise cross-view discrimination term has a worst-case complexity of $\mathcal O(b^2|\mathcal V|^2d)$. PoE fusion, multi-label classification, and active view-label fusion introduce $\mathcal O(b|\mathcal V|d)$, $\mathcal O(b(|\mathcal V|+1)dc)$, and $\mathcal O(b|\mathcal V|c)$ costs, respectively. Therefore, the main additional cost arises from proposal generation and cross-view posterior discrimination during training. }

\subsection{Remarks: On Shared Semantics and View-Label Relevance}
\textbf{Q1: Does the semantic consensus assumption contradict the claim of view complementarity?}
{No. The task-related common semantic assumption is introduced at the representation learning level. It encourages the latent representation inferred from incomplete views to preserve semantic information that is commonly supported by multiple views and relevant to the label space, thereby providing a stable basis for cross-view alignment. This does not imply that different views contribute equally to all labels, nor does it exclude label-specific complementary evidence.}

{At the decision level, V2L retains an individual posterior sample
$z_i^{(v)}$, classifier $g^{(v)}$, and logit vector
$\ell_i^{(v)}$ for each available view. These view-specific branches
allow label-discriminative evidence that is not fully represented
by the shared posterior to contribute through late fusion. The
relevance graph $\mathbf B$ then determines the contribution of each
available view separately for every instance and label. Therefore,
view complementarity in V2L refers to the preservation and adaptive
use of useful predictive differences, rather than encouraging
arbitrary disagreement across views.}

\textbf{Q2: Why do we model instance-wise label-wise view relevance rather than using global view weights?}
Global view weighting assumes that the contribution of each view is fixed across labels and instances, which is overly restrictive for incomplete multi-view multi-label learning. In real scenarios, view importance can vary with both the target label and the input instance. Moreover, solely relying on a fused representation may suppress fine-grained view-specific cues that are useful for particular labels. By estimating instance-wise label-wise view relevance, our model explicitly captures such heterogeneous dependencies and adaptively aggregates view-specific predictions. This enables the model to benefit from both robust shared semantic fusion and flexible label-specific complementary fusion.

\section{Experiments}
\subsection{Experimental Settings}

\subsubsection{Datasets}
Following recent iM3C studies \cite{li2022concise,liu2023dicnet,liu2024attention}, we evaluate the proposed method on five widely-used multi-view multi-label benchmarks: Corel5k \cite{duygulu2002object}, Pascal07 \cite{everingham2009pascal}, ESPGame \cite{von2004labeling}, IAPRTC12 \cite{henning2006iapr}, and MIRFLICKR \cite{huiskes2008mir}. For all datasets, each instance is represented by six visual views, namely GIST, HSV, DenseHue, DenseSIFT, RGB, and LAB. We use the same feature settings and benchmark splits as the existing iM3C literature to ensure fair comparison. Detailed statistics are listed in Table \ref{tab:data}, and we briefly introduce these datasets as follows:
\begin{table}[t]
	\centering
	\caption{Detailed statistics about five multi-view multi-label datasets.}
	\label{tab:data}
	\resizebox{0.45\textwidth}{!}{
		\begin{tabular}{ccccc}
			\toprule[1.2pt]
			Database   & \# Sample  & \# Label & \# View &   \# Label/\#Sample       \\ \midrule
			Corel5k &    4999    &    260    &    6    &    3.40    \\
			Pascal07   &    9963     &    20    &    6    & 1.47 \\
			ESPGame  &    20770     &    268    &    6     &     4.69     \\
			IAPRTC12   &    19627     &    291    &    6     &  5.72   \\
			MIRFLICKR   &    25000     &    38    &    6    &     4.72     \\ 
			\bottomrule[1.2pt]
	\end{tabular}}
\end{table}
\begin{itemize}
\item \textit{Corel5k} is a widely used benchmark for image annotation and multi-label classification, containing 4,999 images and 260 semantic labels. Each image is usually associated with 1 to 5 tags.
\item \textit{Pascal07} is derived from the PASCAL VOC 2007 benchmark and contains 9,963 images with 20 object categories. It is one of the most popular datasets for evaluating visual recognition methods.
\item \textit{ESPGame} consists of 20,770 images collected from an online image-labeling game and is annotated with 268 labels. Each image contains an average of 4.69 semantic labels.
\item \textit{IAPRTC12} is a large-scale benchmark for image annotation and retrieval, containing 19,627 images and 291 categories. Each image can be associated with up to 23 labels.
\item \textit{MIRFLICKR} contains 25,000 images collected from Flickr and includes 38 semantic categories. It provides relatively rich user-generated annotations for multi-label visual understanding.
\end{itemize}

\subsubsection{Construction of incomplete data}
{To simulate the incomplete multi-view missing multi-label setting, we follow the standard protocol used in previous iM3C studies \cite{tan2018incomplete,li2022concise,liu2023dicnet,liu2024partial}. Specifically, for each dataset, we randomly remove a certain proportion of instances on each view while ensuring that every sample has at least one available view. Meanwhile, for each sample, we randomly mask a proportion of positive and negative labels as unknown. Unless otherwise specified, the missing-view ratio and missing-label ratio are both set to 50\%. 
After constructing the incomplete datasets, we randomly split each dataset into training, validation, and test subsets with proportions of 70\%, 15\%, and 15\%, respectively. During training, only the observed views and known labels are used according to the corresponding index matrices. During testing, incomplete views are provided as input, while the evaluation is conducted against the complete ground-truth labels.}

\begin{table*}[t]
	\centering
	\caption{Experimental results of eleven methods on five datasets under 50\% missing views and 50\% missing labels. Standard deviation is listed in the lower right corner. The best result is marked in \textbf{bold}.}
	\label{tab.mainres}
	\resizebox{0.99\textwidth}{!}{
		\begin{tabular}{llcccccccccc>{\columncolor{Gray}}c}
			\toprule[1.1pt]
			\rowcolor{Gray}Data&Metric  &
			CDMM & DM2L & LVSL & iMVWL & NAIM3L & DICNet & DIMC & MSLPP & SIP & QARF & \textbf{Ours} \\
			\midrule
			\multirow{6}[3]{*}{\begin{turn}{90}{Corel5k}\end{turn}}
			&AP $\uparrow$      & 0.354{\tiny0.004} & 0.262{\tiny0.005} & 0.342{\tiny0.004} & 0.283{\tiny0.008} & 0.309{\tiny0.004} & 0.381{\tiny0.004} & 0.353{\tiny0.006} & 0.413{\tiny0.008} & 0.418{\tiny0.009} & 0.411{\tiny0.008} & \textbf{0.426}{\tiny0.011} \\
			&1-HL $\uparrow$    & 0.987{\tiny0.000} & 0.987{\tiny0.000} & 0.987{\tiny0.000} & 0.978{\tiny0.000} & 0.987{\tiny0.000} & 0.988{\tiny0.000} & 0.987{\tiny0.000} & 0.988{\tiny0.000} & 0.988{\tiny0.000} & 0.988{\tiny0.000} & \textbf{0.988}{\tiny0.000} \\
			&1-RL $\uparrow$    & 0.884{\tiny0.003} & 0.843{\tiny0.002} & 0.881{\tiny0.003} & 0.865{\tiny0.005} & 0.878{\tiny0.002} & 0.882{\tiny0.004} & 0.867{\tiny0.001} & 0.901{\tiny0.003} & 0.911{\tiny0.003} & 0.900{\tiny0.004} & \textbf{0.915}{\tiny0.003} \\
			&AUC $\uparrow$     & 0.888{\tiny0.003} & 0.845{\tiny0.002} & 0.884{\tiny0.003} & 0.868{\tiny0.005} & 0.881{\tiny0.002} & 0.884{\tiny0.004} & 0.870{\tiny0.001} & 0.903{\tiny0.004} & 0.913{\tiny0.003} & 0.902{\tiny0.004} & \textbf{0.916}{\tiny0.003} \\
			&1-OE $\uparrow$    & 0.410{\tiny0.007} & 0.295{\tiny0.014} & 0.391{\tiny0.009} & 0.311{\tiny0.015} & 0.350{\tiny0.009} & 0.468{\tiny0.007} & 0.422{\tiny0.015} & 0.485{\tiny0.010} & 0.489{\tiny0.016} & 0.492{\tiny0.013} & \textbf{0.494}{\tiny0.013} \\
			&1-Cov $\uparrow$   & 0.723{\tiny0.007} & 0.647{\tiny0.005} & 0.718{\tiny0.006} & 0.702{\tiny0.008} & 0.725{\tiny0.005} & 0.727{\tiny0.011} & 0.684{\tiny0.011} & 0.766{\tiny0.009} & 0.787{\tiny0.009} & 0.761{\tiny0.009} & \textbf{0.793}{\tiny0.008} \\
			\midrule\midrule
			\multirow{6}[3]{*}{\begin{turn}{90}{Pascal07}\end{turn}}
			&AP $\uparrow$      & 0.508{\tiny0.005} & 0.471{\tiny0.008} & 0.504{\tiny0.005} & 0.437{\tiny0.018} & 0.488{\tiny0.003} & 0.505{\tiny0.012} & 0.532{\tiny0.002} & 0.544{\tiny0.010} & 0.555{\tiny0.010} & 0.556{\tiny0.007} & \textbf{0.562}{\tiny0.008} \\
			&1-HL $\uparrow$    & 0.931{\tiny0.001} & 0.928{\tiny0.001} & 0.930{\tiny0.000} & 0.882{\tiny0.004} & 0.928{\tiny0.001} & 0.929{\tiny0.001} & 0.931{\tiny0.001} & 0.932{\tiny0.001} & 0.931{\tiny0.001} & 0.931{\tiny0.001} & \textbf{0.934}{\tiny0.001} \\
			&1-RL $\uparrow$    & 0.812{\tiny0.004} & 0.761{\tiny0.005} & 0.806{\tiny0.003} & 0.736{\tiny0.015} & 0.783{\tiny0.001} & 0.783{\tiny0.008} & 0.813{\tiny0.000} & 0.819{\tiny0.006} & 0.830{\tiny0.004} & 0.828{\tiny0.004} & \textbf{0.836}{\tiny0.004} \\
			&AUC $\uparrow$     & 0.838{\tiny0.003} & 0.779{\tiny0.004} & 0.832{\tiny0.002} & 0.767{\tiny0.015} & 0.811{\tiny0.001} & 0.809{\tiny0.006} & 0.833{\tiny0.002} & 0.841{\tiny0.004} & 0.850{\tiny0.005} & 0.847{\tiny0.004} & \textbf{0.858}{\tiny0.003} \\
			&1-OE $\uparrow$    & 0.419{\tiny0.008} & 0.420{\tiny0.011} & 0.419{\tiny0.008} & 0.362{\tiny0.023} & 0.421{\tiny0.006} & 0.427{\tiny0.015} & 0.456{\tiny0.011} & 0.466{\tiny0.014} & 0.464{\tiny0.018} & {0.471}{\tiny0.011} & \textbf{0.473}{\tiny0.010} \\
			&1-Cov $\uparrow$   & 0.759{\tiny0.003} & 0.692{\tiny0.004} & 0.751{\tiny0.003} & 0.677{\tiny0.015} & 0.727{\tiny0.002} & 0.731{\tiny0.006} & 0.769{\tiny0.007} & 0.771{\tiny0.003} & 0.783{\tiny0.006} & 0.780{\tiny0.005} & \textbf{0.793}{\tiny0.004} \\
			\midrule\midrule
			\multirow{6}[3]{*}{\begin{turn}{90}{ESPGame}\end{turn}}
			&AP $\uparrow$      & 0.289{\tiny0.003} & 0.212{\tiny0.002} & 0.285{\tiny0.003} & 0.244{\tiny0.005} & 0.246{\tiny0.002} & 0.297{\tiny0.002} & 0.287{\tiny0.002} & 0.310{\tiny0.004} & 0.311{\tiny0.004} & 0.307{\tiny0.004} & \textbf{0.321}{\tiny0.005} \\
			&1-HL $\uparrow$    & 0.983{\tiny0.000} & 0.982{\tiny0.000} & 0.983{\tiny0.000} & 0.972{\tiny0.000} & 0.983{\tiny0.000} & 0.983{\tiny0.000} & 0.983{\tiny0.000} & 0.983{\tiny0.000} & 0.983{\tiny0.000} & 0.983{\tiny0.000} & \textbf{0.983}{\tiny0.000} \\
			&1-RL $\uparrow$    & 0.832{\tiny0.001} & 0.781{\tiny0.001} & 0.829{\tiny0.001} & 0.808{\tiny0.002} & 0.818{\tiny0.002} & 0.832{\tiny0.001} & 0.821{\tiny0.000} & 0.843{\tiny0.002} & 0.849{\tiny0.002} & 0.842{\tiny0.001} & \textbf{0.853}{\tiny0.002} \\
			&AUC $\uparrow$     & 0.836{\tiny0.001} & 0.785{\tiny0.001} & 0.833{\tiny0.002} & 0.813{\tiny0.002} & 0.824{\tiny0.002} & 0.836{\tiny0.001} & 0.826{\tiny0.000} & 0.847{\tiny0.002} & 0.853{\tiny0.002} & 0.846{\tiny0.001} & \textbf{0.857}{\tiny0.002} \\
			&1-OE $\uparrow$    & 0.396{\tiny0.005} & 0.294{\tiny0.006} & 0.389{\tiny0.004} & 0.343{\tiny0.013} & 0.339{\tiny0.003} & 0.439{\tiny0.007} & 0.435{\tiny0.009} & 0.457{\tiny0.012} & 0.455{\tiny0.007} & 0.448{\tiny0.009} & \textbf{0.467}{\tiny0.011} \\
			&1-Cov $\uparrow$   & 0.574{\tiny0.004} & 0.488{\tiny0.003} & 0.567{\tiny0.005} & 0.548{\tiny0.004} & 0.571{\tiny0.003} & 0.593{\tiny0.003} & 0.562{\tiny0.004} & 0.622{\tiny0.005} & 0.628{\tiny0.005} & \textbf{0.651}{\tiny0.004} & {0.638}{\tiny0.005} \\
			\midrule\midrule
			\multirow{6}[3]{*}{\begin{turn}{90}{IAPRTC12}\end{turn}}
			&AP $\uparrow$      & 0.305{\tiny0.004} & 0.234{\tiny0.003} & 0.304{\tiny0.004} & 0.237{\tiny0.003} & 0.261{\tiny0.001} & 0.323{\tiny0.001} & 0.308{\tiny0.001} & {0.340}{\tiny0.005} & 0.331{\tiny0.006} & 0.338{\tiny0.004} & \textbf{0.349}{\tiny0.005} \\
			&1-HL $\uparrow$    & 0.981{\tiny0.000} & 0.980{\tiny0.000} & 0.981{\tiny0.000} & 0.969{\tiny0.000} & 0.980{\tiny0.000} & 0.981{\tiny0.000} & 0.980{\tiny0.000} & 0.981{\tiny0.000} & 0.980{\tiny0.000} & 0.981{\tiny0.000} & \textbf{0.981}{\tiny0.000} \\
			&1-RL $\uparrow$    & 0.862{\tiny0.002} & 0.823{\tiny0.002} & 0.861{\tiny0.002} & 0.833{\tiny0.002} & 0.848{\tiny0.001} & 0.873{\tiny0.001} & 0.864{\tiny0.000} & 0.882{\tiny0.002} & 0.885{\tiny0.003} & 0.882{\tiny0.002} & \textbf{0.893}{\tiny0.002} \\
			&AUC $\uparrow$     & 0.864{\tiny0.002} & 0.825{\tiny0.001} & 0.863{\tiny0.001} & 0.835{\tiny0.001} & 0.850{\tiny0.001} & 0.874{\tiny0.000} & 0.864{\tiny0.000} & 0.883{\tiny0.002} & 0.886{\tiny0.002} & 0.883{\tiny0.002} & \textbf{0.893}{\tiny0.002} \\
			&1-OE $\uparrow$    & 0.432{\tiny0.008} & 0.340{\tiny0.006} & 0.429{\tiny0.009} & 0.352{\tiny0.008} & 0.390{\tiny0.005} & 0.468{\tiny0.002} & 0.431{\tiny0.006} & 0.474{\tiny0.008} & 0.463{\tiny0.009} & 0.471{\tiny0.011} & \textbf{0.482}{\tiny0.001} \\
			&1-Cov $\uparrow$   & 0.597{\tiny0.004} & 0.529{\tiny0.004} & 0.597{\tiny0.004} & 0.564{\tiny0.005} & 0.592{\tiny0.004} & 0.649{\tiny0.001} & 0.597{\tiny0.004} & 0.672{\tiny0.006} & 0.675{\tiny0.007} & 0.668{\tiny0.005} & \textbf{0.694}{\tiny0.006} \\
			\midrule\midrule
			\multirow{6}[3]{*}{\begin{turn}{90}{MIRFLICKR}\end{turn}}
			&AP $\uparrow$      & 0.570{\tiny0.002} & 0.514{\tiny0.006} & 0.553{\tiny0.002} & 0.490{\tiny0.012} & 0.551{\tiny0.002} & 0.589{\tiny0.005} & 0.602{\tiny0.002} & 0.615{\tiny0.004} & 0.614{\tiny0.004} & 0.616{\tiny0.004} & \textbf{0.621}{\tiny0.003} \\
			&1-HL $\uparrow$    & 0.886{\tiny0.001} & 0.878{\tiny0.001} & 0.885{\tiny0.001} & 0.839{\tiny0.002} & 0.882{\tiny0.001} & 0.888{\tiny0.002} & 0.888{\tiny0.000} & 0.892{\tiny0.001} & 0.891{\tiny0.001} & 0.892{\tiny0.001} & \textbf{0.895}{\tiny0.001} \\
			&1-RL $\uparrow$    & 0.856{\tiny0.001} & 0.831{\tiny0.003} & 0.856{\tiny0.001} & 0.803{\tiny0.008} & 0.844{\tiny0.001} & 0.863{\tiny0.004} & 0.865{\tiny0.001} & 0.879{\tiny0.002} & 0.877{\tiny0.002} & 0.880{\tiny0.003} & \textbf{0.881}{\tiny0.003} \\
			&AUC $\uparrow$     & 0.846{\tiny0.001} & 0.828{\tiny0.003} & 0.844{\tiny0.001} & 0.787{\tiny0.012} & 0.837{\tiny0.001} & 0.849{\tiny0.004} & 0.852{\tiny0.001} & 0.865{\tiny0.002} & 0.860{\tiny0.003} & 0.866{\tiny0.002} & \textbf{0.872}{\tiny0.002} \\
			&1-OE $\uparrow$    & 0.631{\tiny0.004} & 0.510{\tiny0.008} & 0.607{\tiny0.004} & 0.511{\tiny0.022} & 0.585{\tiny0.003} & 0.637{\tiny0.007} & 0.647{\tiny0.007} & 0.667{\tiny0.007} & 0.662{\tiny0.008} & 0.662{\tiny0.008} & \textbf{0.673}{\tiny0.007} \\
			&1-Cov $\uparrow$   & 0.640{\tiny0.001} & 0.604{\tiny0.005} & 0.636{\tiny0.001} & 0.572{\tiny0.013} & 0.631{\tiny0.002} & 0.652{\tiny0.007} & 0.661{\tiny0.003} & 0.679{\tiny0.003} & 0.678{\tiny0.003} & 0.683{\tiny0.003} & \textbf{0.683}{\tiny0.003} \\
			\bottomrule[1.1pt]
	\end{tabular}}
	\vspace{-0.5cm}
\end{table*}
\subsubsection{Competitors}
To evaluate the performance of our proposed V2L, we compare our method with representative methods from three related categories, including complete multi-view multi-label methods CDMM \cite{zhao2021consistency} and LVSL \cite{zhao2022non}, incomplete multi-label learning method DM2L \cite{ma2021expand}, and dedicated iM3C methods iMVWL \cite{tan2018incomplete}, NAIM3L \cite{li2022concise}, DICNet \cite{liu2023dicnet}, DIMC \cite{wen2023deep}, MSLPP \cite{long2024multi}, SIP \cite{liu2024partial}, and QARF \cite{lu2025partial}. For methods that are not originally designed for simultaneous view and label incompleteness, we follow the standard adaptation protocol used in prior work \cite{zhao2022non,li2022concise}: missing views are imputed by the corresponding average feature, and unobserved labels are excluded from the supervised loss. For the fairness of comparison, we use the released code and tune the hyper-parameters according to the recommended settings.

\subsubsection{Evaluation Metrics}
In order to be consistent with the existing work, we report six standard multi-label metrics, including Average Precision (AP), Hamming Loss (HL), Ranking Loss (RL), Area Under the ROC Curve (AUC), One-Error (OE), and Coverage (Cov). To keep the direction of all criteria consistent, we report 1-HL, 1-RL, 1-OE, and 1-Cov together with AP and AUC, so that larger values indicate better performance in every case. Following the common protocol in multi-label evaluation, the final results are reported as the mean and standard deviation over 10 folds.

\subsubsection{Implementation Details}
Our V2L is implemented in PyTorch. Unless otherwise stated, the latent dimension is set to 512, the batch size is set to 128, and the model is optimized by Adam with learning rate $1\times 10^{-3}$. The total training epoch number is 100, where the first 10 epochs are used as a warm-up stage before activating the view-label relevance supervision term. The trade-off coefficients are selected on the validation set from the candidate ranges used in training. 

\subsection{Experimental Results and Analysis}
\begin{figure*}[t!]
	\centering
	\subfloat[Corel5k]{
		\includegraphics[width=0.32\textwidth]{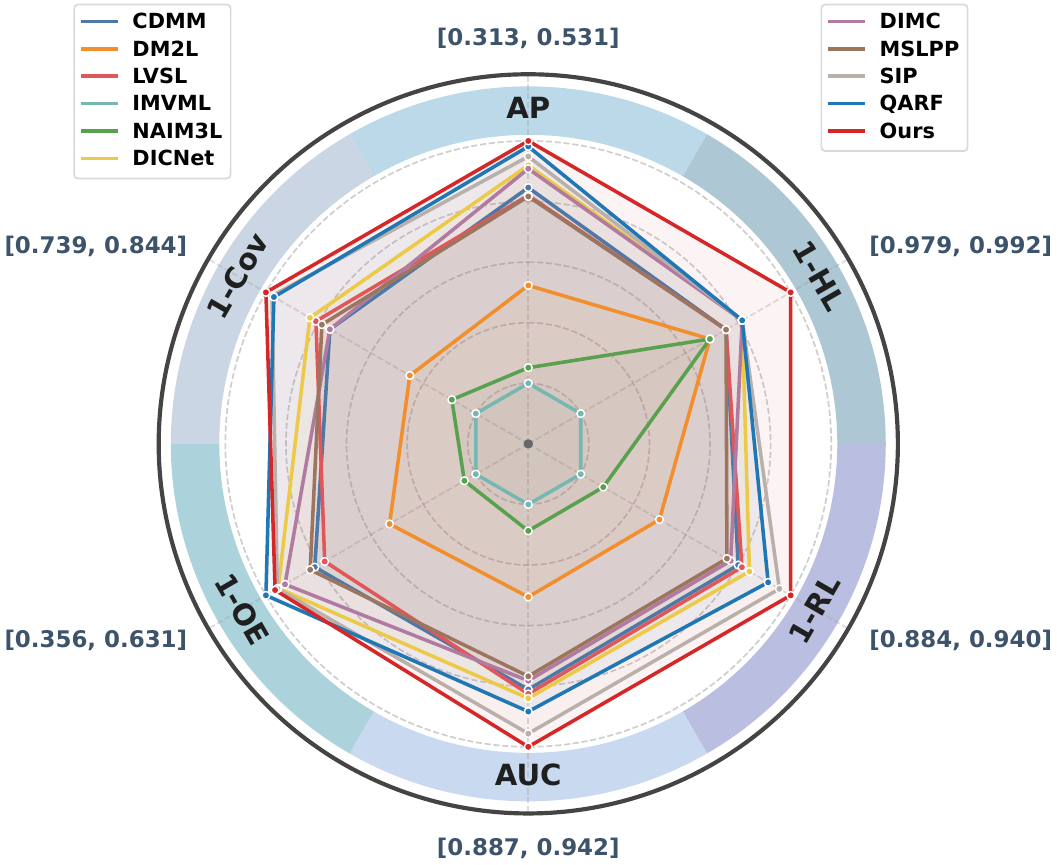}
	}
	\subfloat[Pascal07]{
		
		\includegraphics[width=0.32\textwidth]{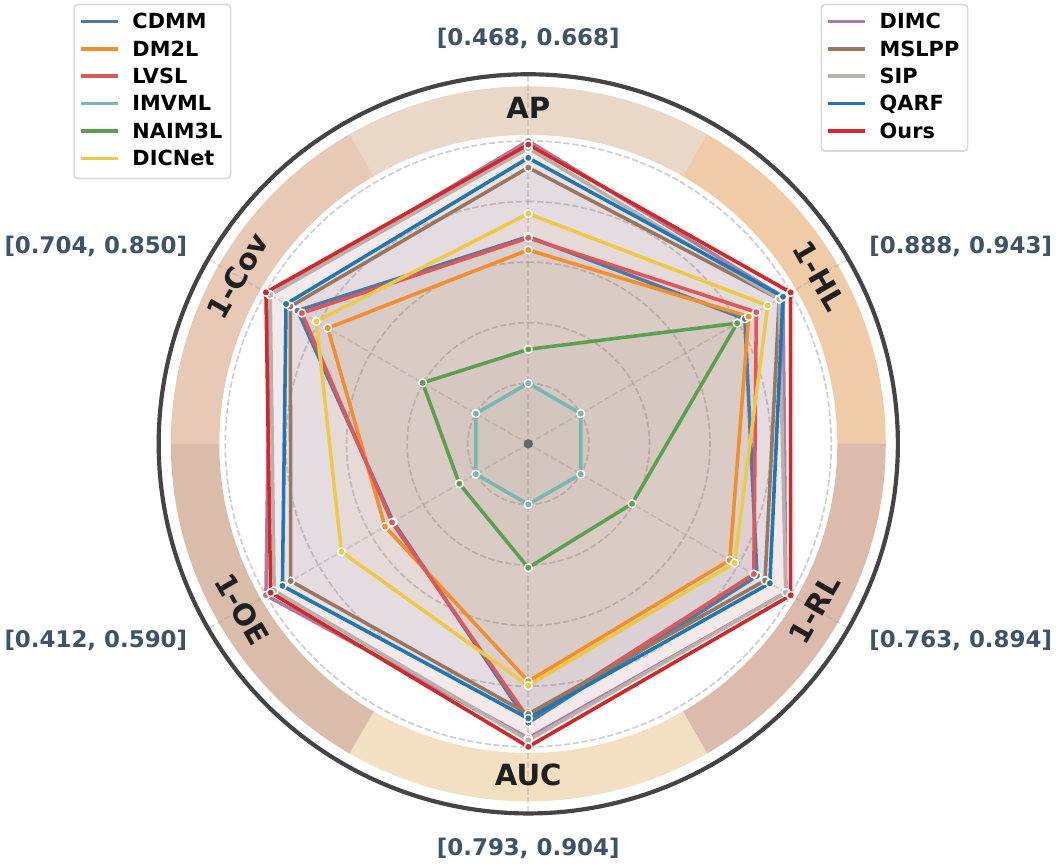}
	}
	\subfloat[ESPGame]{
		
		\includegraphics[width=0.32\textwidth]{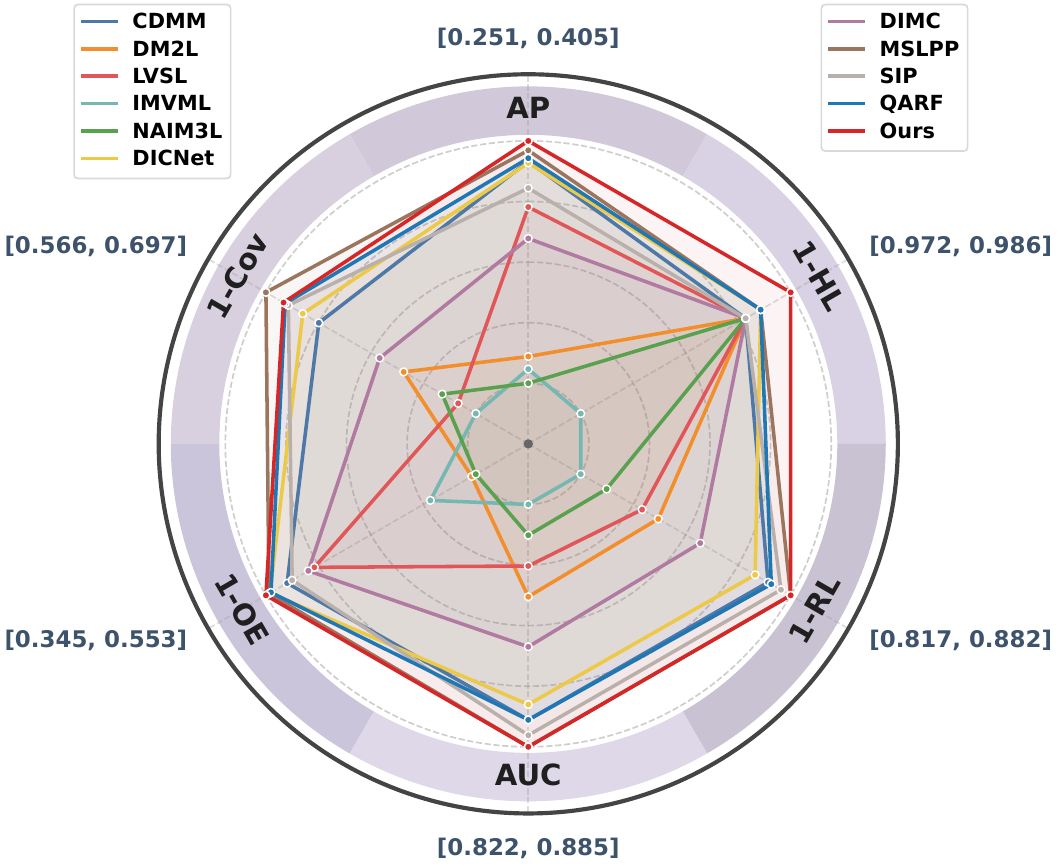}
	}
	\quad
	\subfloat[IAPRTC12]{
		
		\includegraphics[width=0.32\textwidth]{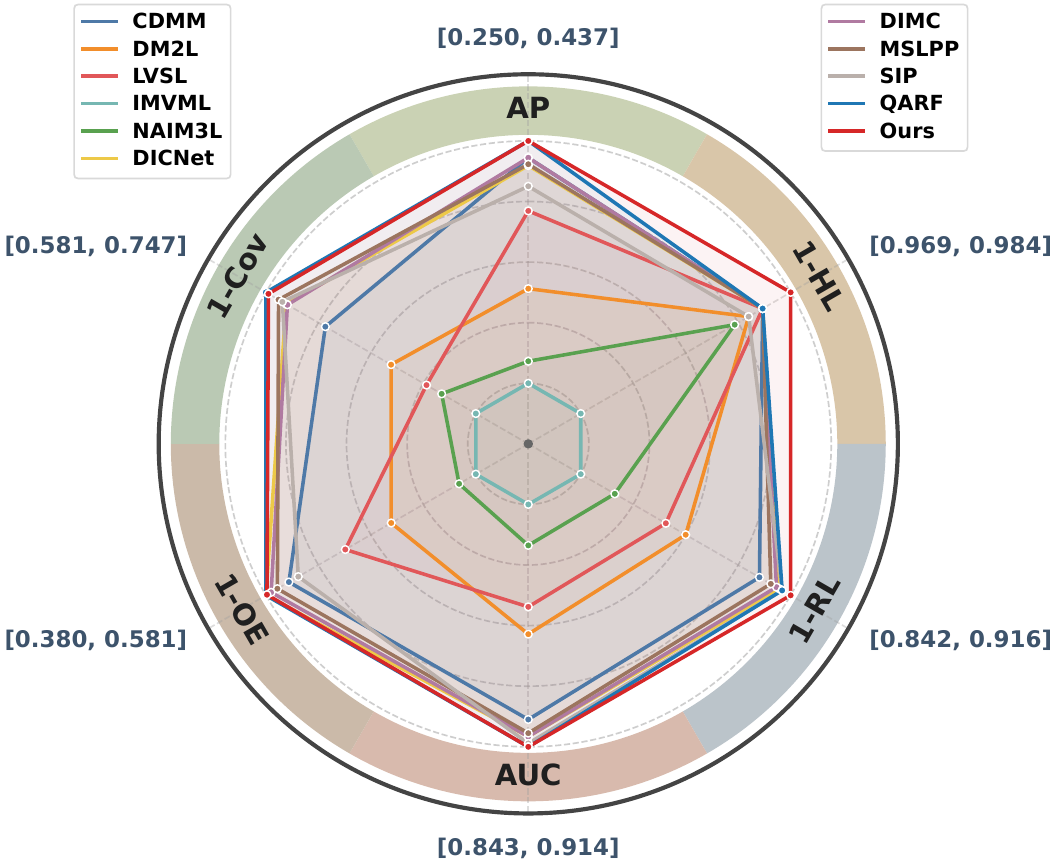}
	}
	\subfloat[MIRFLICKR]{
		
		\includegraphics[width=0.32\textwidth]{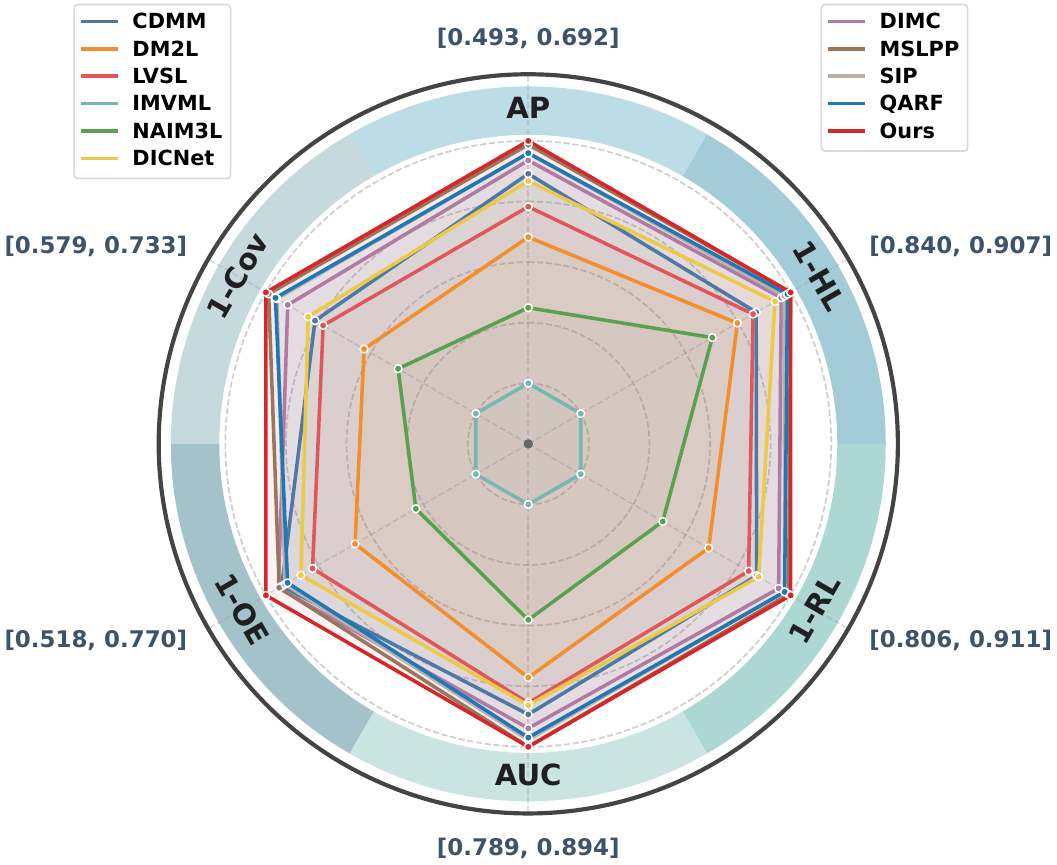}
	}
	
	\caption{Experimental results of eleven methods on the five full datasets without any missing views or labels. The center of the radar map shows the worst results and the vertexes mean the best results on the six metrics.}
	\label{fig.res0}
	\vspace{-0.2cm}
\end{figure*}
\begin{figure}[t!]
	\centering
	\subfloat[MIRFLICKR]{
		\includegraphics[width=0.45\linewidth]{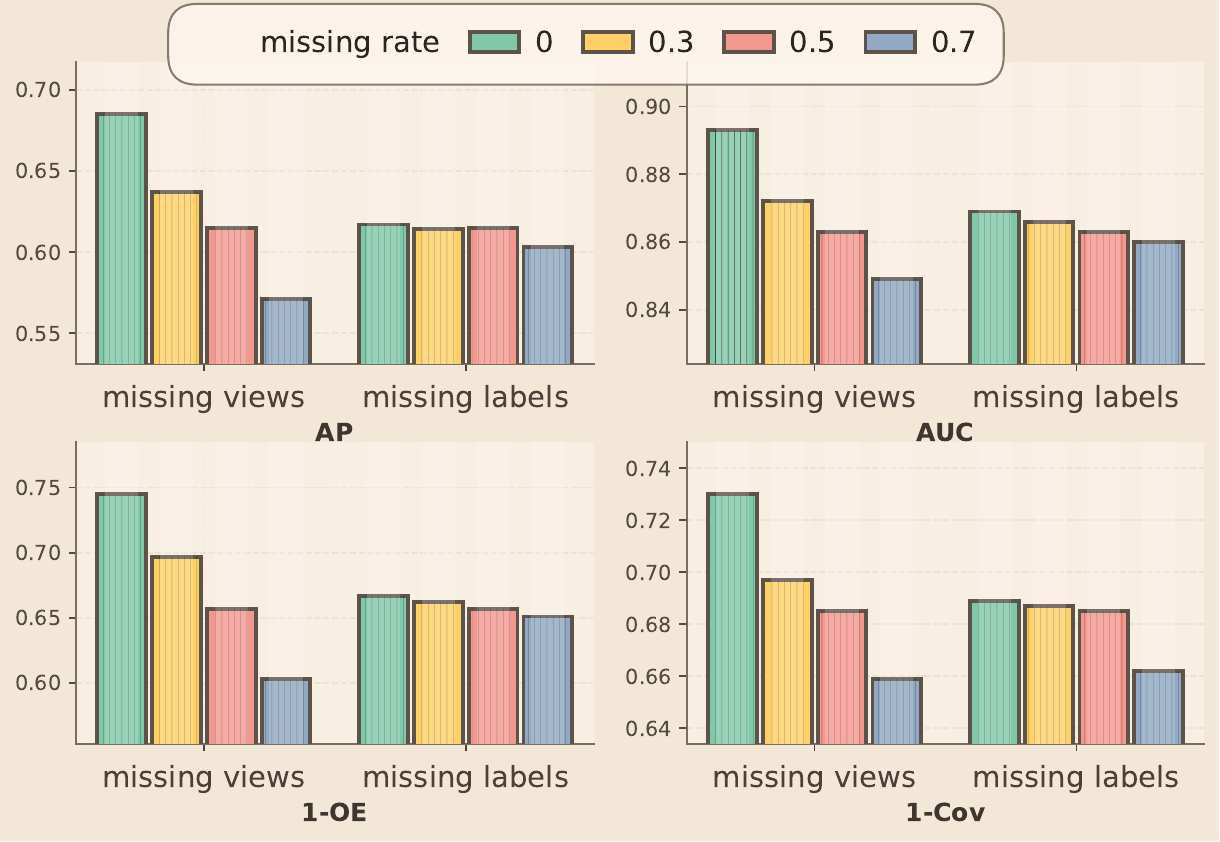}
	}
	\subfloat[Pascal07]{
		\includegraphics[width=0.45\linewidth]{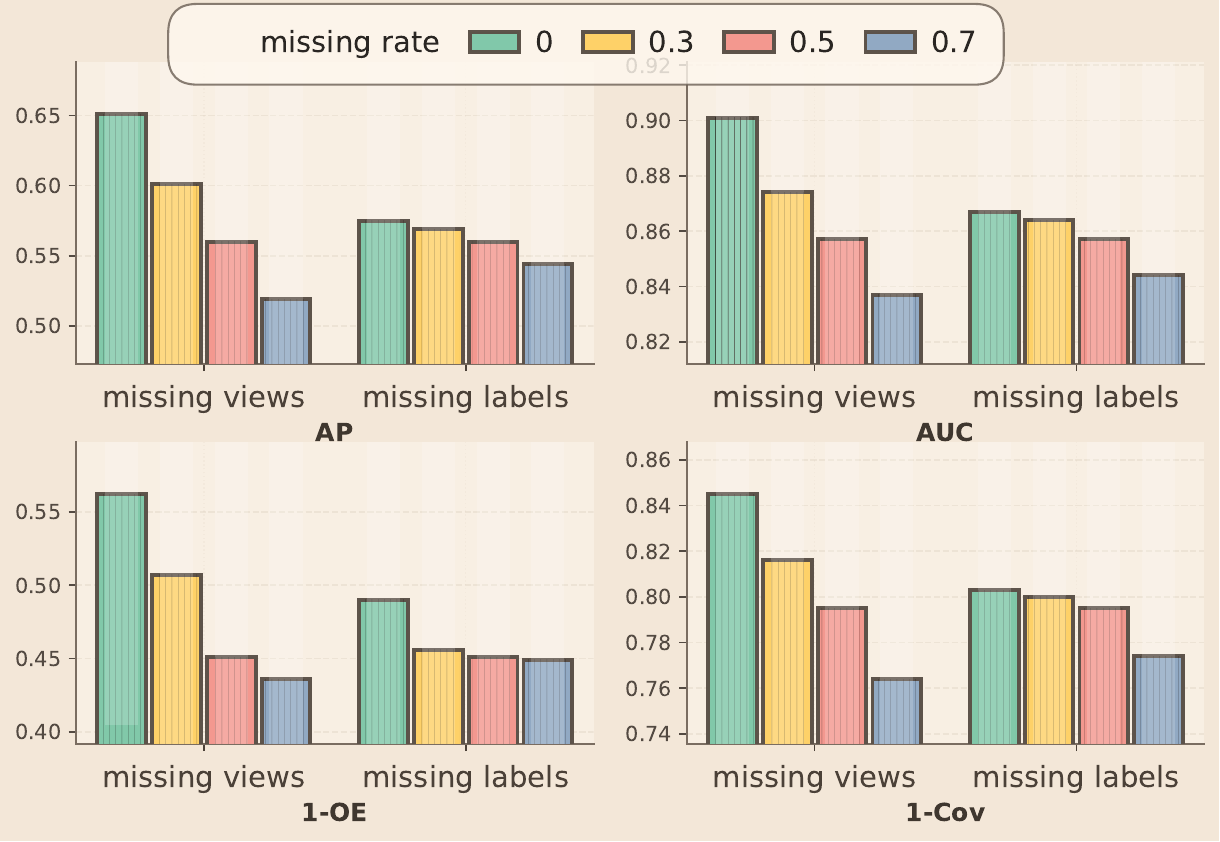}
	}
	\caption{Experimental results under different missing rates on the MIRFLICKR and Pascal07 datasets. We respectively fix the view/label missing rate at 0.5, and change the other one to 0, 0.3, 0.5, and 0.7.}
	\label{fig.exp_missing_perf}
	\vspace{-0.3cm}
\end{figure}
\subsubsection{Experiments on Incomplete Settings}
To verify the performance of our V2L, we first report the main comparison results under the standard doubly-incomplete setting, where both the view missing rate and the label missing rate are fixed at 50\%. The quantitative results are summarized in Table \ref{tab.mainres}. The advantage of V2L is not limited to a single evaluation criterion. Improvements on AP and AUC indicate that the learned representation provides reliable global ranking of labels, while gains on 1-OE and 1-Cov suggest that the adaptive decision branch helps identify the most confident labels more accurately. This is consistent with the design of V2L: the semantic encoding branch stabilizes the latent space under missing views, whereas the view-label relevance branch refines label-wise decisions by exploiting view-specific evidence. 

These results validate the effectiveness of the two core components in our framework. The source-perturbation invariant encoding preserves shared task-relevant semantics from incomplete observations, which directly benefits holistic metrics such as AP and AUC. Meanwhile, the active view-label relevance modeling performs instance-wise and label-aware predictive fusion, which further improves ranking-oriented criteria such as $1$-RL, $1$-OE, and $1$-Cov. Compared with existing methods that mainly emphasize either shared latent learning or global late fusion, V2L better couples cross-view semantic consistency with label-sensitive decision aggregation, thereby producing more reliable predictions in the iM3C setting.
\subsubsection{Experiments on Complete and Varying Missing Settings}
To further examine whether the superiority of V2L is limited to incomplete scenarios, we additionally evaluate all methods on the fully observed setting without missing views or labels. The radar plots in Fig. \ref{fig.res0} show that V2L still remains among the top-performing methods across different datasets and evaluation metrics. This observation is important because it indicates that our model does not gain its advantage merely from specialized missing-data handling. Instead, the learned shared representation and adaptive view-label fusion strategy also generalize well to the ideal complete-data regime, showing that V2L is a unified multi-view multi-label learner rather than a method tailored only for degraded inputs.

In addition, to investigate the influence of different missing patterns, we visualize the results under varying view and label missing rates in Fig. \ref{fig.exp_missing_perf}. Specifically, we fix one missing rate at 50\% and change the other from 0 to 0.7. A clear performance decline can be observed as either type of missingness increases, which agrees with the intuition that both missing observations and missing supervision reduce the available discriminative information. At the same time, the degradation caused by increasing the view missing rate is generally more pronounced than that caused by increasing the label missing rate. This suggests that, although incomplete supervision is harmful, the loss of observed views more directly weakens cross-view semantic aggregation and reduces the complementary evidence available for label prediction. Overall, these results confirm that V2L maintains favorable performance over a broad range of missing conditions and degrades in a stable rather than abrupt manner.

\begin{figure*}[t!]
	\centering
	\subfloat[Instance A in Pascal07]{
		\includegraphics[width=0.49\textwidth]{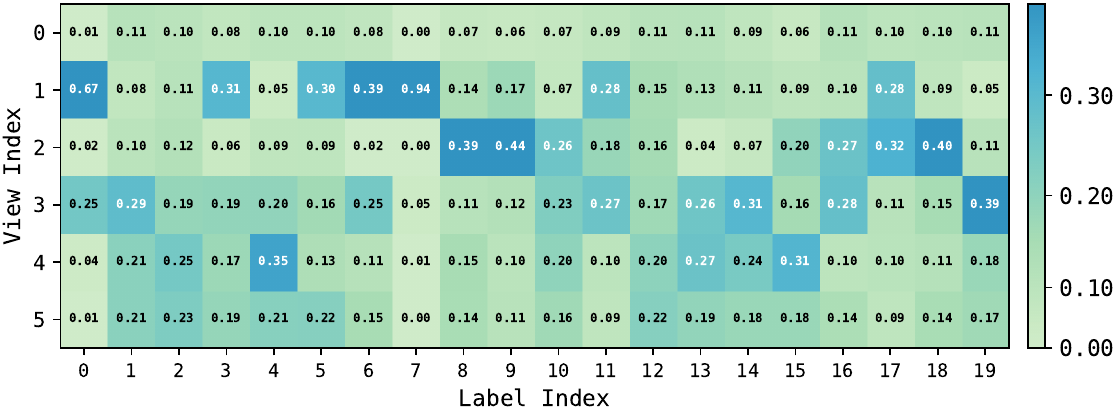}
	}
	\subfloat[Instance B in Pascal07]{
			\includegraphics[width=0.49\textwidth]{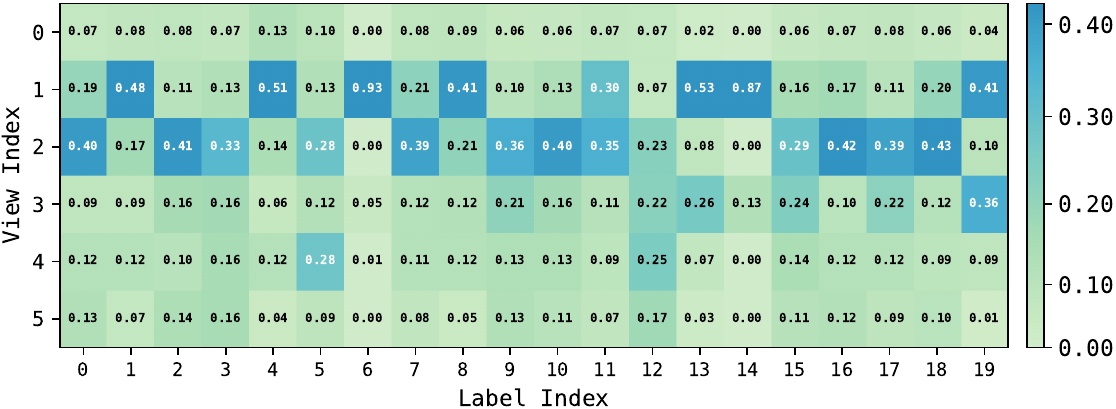}
		}
	\quad
	\subfloat[Instance A in MIRFLICKR]{
		\includegraphics[width=0.49\textwidth]{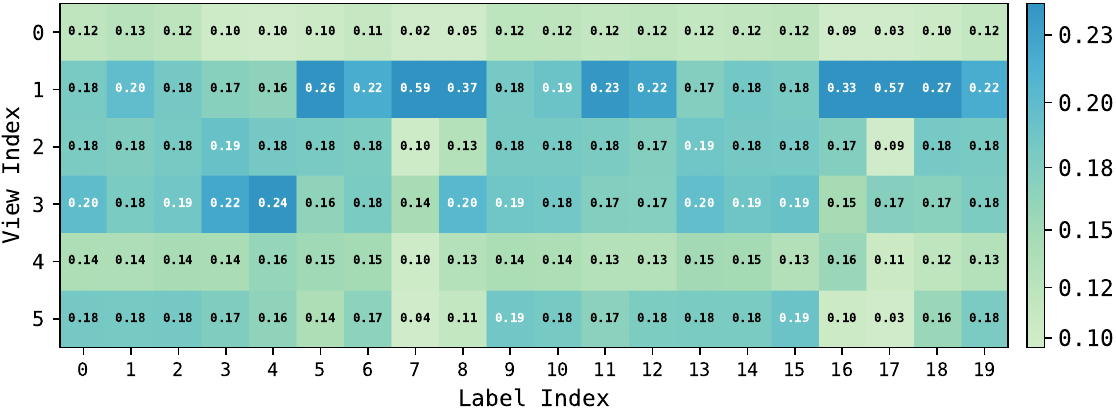}
	}
	\subfloat[Instance B in MIRFLICKR]{
			\includegraphics[width=0.49\textwidth]{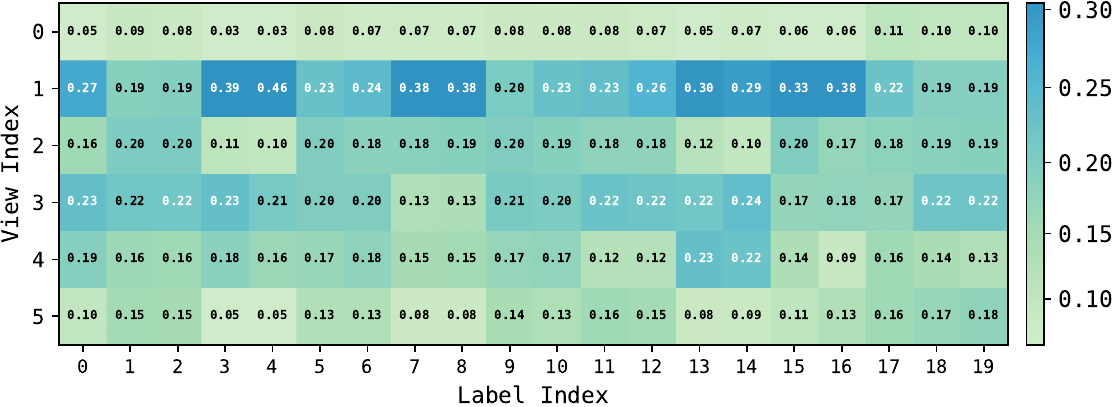}
		}
	\caption{Illustration of view-label heterogeneity modeling. We visualize the heat map of view-label relevance graph of two random instances from the Pascal07 and MIRFLICKR datasets (Only the first 20 tags of the Pascal07 dataset are displayed due to space limitations).}
	\label{fig.exp_view_label}
	\vspace{-0.2cm}
\end{figure*}

\subsection{Study of View-Label Heterogeneity}
In this section, to verify the effectiveness of our view-label heterogeneity modeling strategy, we visualize the view-label relevance graph predicted by the proposed active perception network in Fig.~\ref{fig.exp_view_label}. Specifically, we conduct experiments on the complete Pascal07 and MIRFLICKR datasets, and randomly select two instances from each dataset for case study. For each instance, we plot the heat map of the correlations between its six views and the first 20 labels. These visualizations provide a direct way to examine whether the model indeed learns instance-wise and label-aware view contributions, rather than relying on static view fusion.

Three observations can be drawn from Fig.~\ref{fig.exp_view_label}: (1) The heat maps of different instances are clearly different even within the same dataset, indicating that the contribution pattern of views is instance-dependent rather than globally fixed. This observation supports our motivation that view relevance should be dynamically evaluated for each instance. (2) For any instance, labels exhibit substantially different dependencies over the six views. In other words, the contribution of the same view to the prediction of different labels varies. This phenomenon directly confirms the necessity of modeling view-label heterogeneity instead of assigning identical fusion weights to all labels. (3) From a complete view perspective, the imbalance among views is also clearly observable. For example, we can intuitively find that view 0 shows relatively weak discriminative ability on both datasets, while some other views contribute much more strongly to the final prediction. Compared with MIRFLICKR, such an imbalance is even more pronounced on Pascal07, where the dominant and weak views are more distinctly separated. These observations further justify the necessity of our composite predictive strategy, i.e., the mid-level fusion captures shared semantics, while the late-fusion adaptively reweights view-specific evidence for each label. Therefore, the visual results not only validate the existence of view-label heterogeneity, but also demonstrate why an instance-wise label-aware post-fusion mechanism is essential for effective multi-view multi-label prediction.

{The low relevance of some views can be explained by differences in feature discriminability and information redundancy. For example, view~0 corresponds to GIST, which mainly captures global spatial structure and may be less informative than local appearance, color, or texture features for object- and tag-oriented labels. Moreover, $\mathbf B_{i,j,v}$ measures the normalized relative contribution among available views rather than absolute usefulness. Thus, a view may receive a low weight when its evidence is less reliable or redundant with that of stronger views, without being entirely uninformative.}
\begin{figure}[t!]
	\centering
	\subfloat[MIRFLICKR]{
		\includegraphics[width=0.24\textwidth]{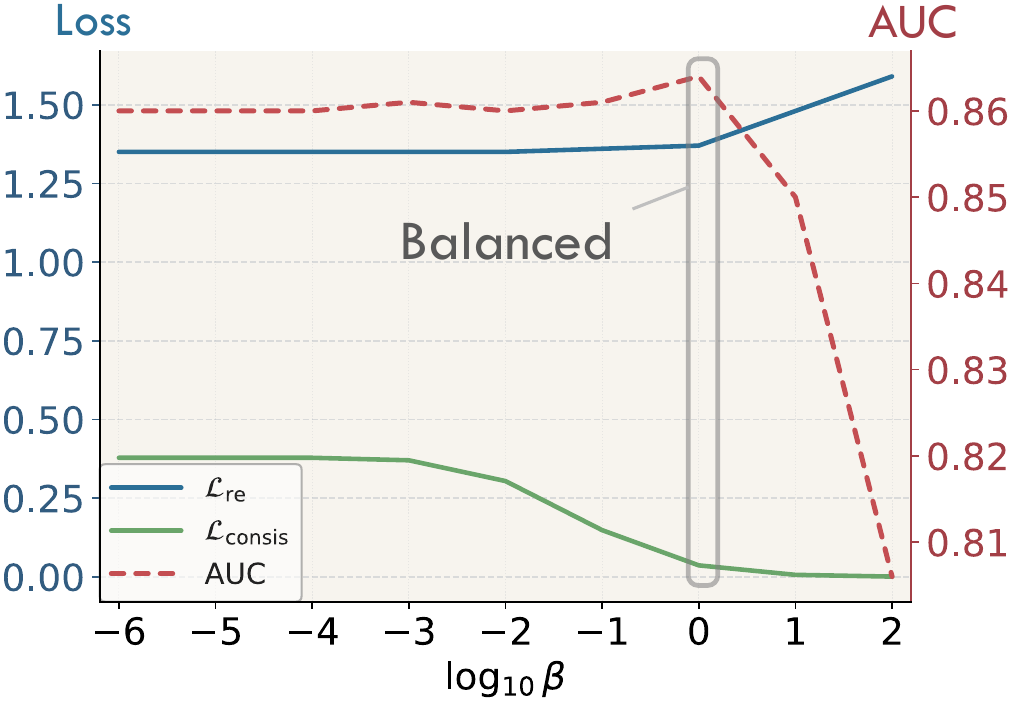}
	}
	\subfloat[Pascal07]{
		\includegraphics[width=0.24\textwidth]{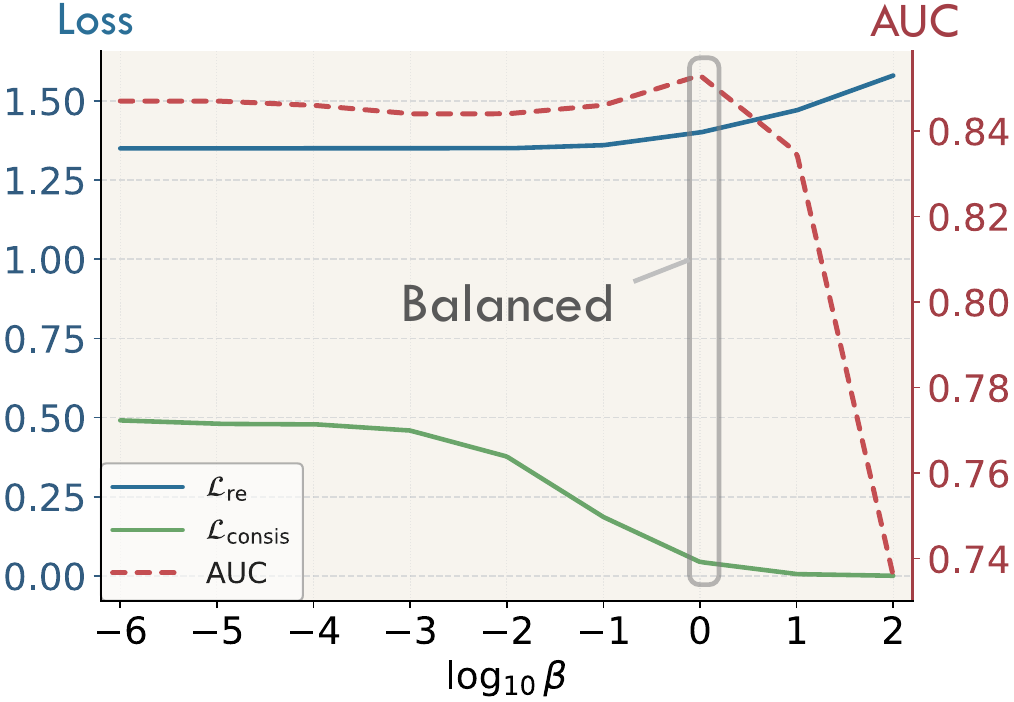}
	}

	\caption{Impact of different information-balance parameter $\beta$ on loss $\mathcal{L}_{\mathrm{re}}$, $\mathcal{L}_{\mathrm{consis}}$, and AUC.}
	\label{fig.exp_ib}
	\vspace{-0.2cm}
\end{figure}
\subsection{Study of Information Bottleneck Design}
In this subsection, we further analyze the self-supervised representation learning part in the overall objective in Eq.~(\ref{eq:overall_obj}), namely the first two terms that are instantiated by the information bottleneck loss $\mathcal{L}_{\mathrm{ib}}$ in Eq.~(\ref{eq:lib_pami}). Recall that $\mathcal{L}_{\mathrm{ib}}=\mathcal{L}_{\mathrm{re}}+\beta\mathcal{L}_{\mathrm{consis}}$, where $\mathcal{L}_{\mathrm{re}}$ encourages each view-specific latent variable to preserve valid information for reconstruction, while $\mathcal{L}_{\mathrm{consis}}$ suppresses view-private redundancy and promotes cross-view semantic consistency. Therefore, the key issue is how to choose the trade-off coefficient $\beta$ so that the learned representation remains both sufficiently informative and compact.

{To study this balance, we vary $\beta$ and visualize the corresponding changes of $\mathcal{L}_{\mathrm{re}}$, $\mathcal{L}_{\mathrm{consis}}$, and AUC on MIRFLICKR and Pascal07 datasets, as shown in Fig.~\ref{fig.exp_ib}. A clear trade-off can be observed: When $\beta$ is too small, the reconstruction term dominates the optimization, so the model tends to retain more view-specific details. Although this behavior keeps $\mathcal{L}_{\mathrm{re}}$ low, the consistency regularization is not strong enough to sufficiently remove non-shared information across views, and the final AUC is therefore suboptimal. In contrast, when $\beta$ becomes too large, the model overemphasizes semantic compression and aggressively minimizes $\mathcal{L}_{\mathrm{consis}}$. This causes $\mathcal{L}_{\mathrm{re}}$ to increase noticeably, indicating that part of the view-valid information useful for reconstruction has been discarded, which also leads to a decline in predictive performance.}

{The best results are obtained in the middle range of $\beta$, where the two objectives reach a favorable balance (In Fig. \ref{fig.exp_ib}, we can see the best $\beta=1e0$). At this stage, the model neither overfits to view-specific redundancy nor over-compresses the latent representation, preserving enough valid information for latent encoding while enforcing sufficient semantic consistency across views. These observations verify that the effectiveness of our information bottleneck design does not come from reconstruction or consistency alone, but from their coordinated interaction. In other words, the two terms of $\mathcal{L}_{\mathrm{ib}}$ indeed play complementary roles in encoding both semantically aligned and informative representations for downstream multi-label prediction.}

\begin{table*}[t!]
\caption{Ablation results with respect to loss functions on Corel5k and Pascal07 datasets under 50\% missing views and 50\% missing labels. `w/o' means `without'.}
\label{tab:abl_loss}
\footnotesize
\centering
\resizebox{0.99\textwidth}{!}{
\rowcolors{3}{LightGray}{white}
\begin{tabular}{c|cccccc|cccccc}
\toprule[1.1pt]
\multirow{2}{*}{Method} & \multicolumn{6}{c|}{Corel5k} & \multicolumn{6}{c}{Pascal07} \\
& AP & 1-HL & 1-RL & AUC & 1-OE & 1-Cov & AP & 1-HL & 1-RL & AUC & 1-OE & 1-Cov \\
\midrule
V2L w/o $\mathcal{L}_{\mathrm{post}}$ $\mathcal{L}_{\mathrm{ib}}$ $\mathcal{L}_{\mathrm{rele}}$ & 0.352 & 0.986 & 0.864 & 0.867 & 0.428 & 0.702 & 0.527 & 0.926 & 0.815 & 0.835 & 0.426 & 0.767 \\
V2L w/o $\mathcal{L}_{\mathrm{ib}}$ $\mathcal{L}_{\mathrm{rele}}$ & 0.374 & 0.987 & 0.891 & 0.893 & 0.441 & 0.752 & 0.534 & 0.929 & 0.824 & 0.844 & 0.427 & 0.777 \\
V2L w/o $\mathcal{L}_{\mathrm{post}}$ $\mathcal{L}_{\mathrm{rele}}$ & 0.390 & 0.986 & 0.898 & 0.900 & 0.464 & 0.762 & 0.541 & 0.928 & 0.827 & 0.845 & 0.444 & 0.780 \\
V2L w/o $\mathcal{L}_{\mathrm{post}}$ $\mathcal{L}_{\mathrm{ib}}$ & 0.371 & 0.987 & 0.865 & 0.867 & 0.463 & 0.700 & 0.526 & 0.926 & 0.813 & 0.833 & 0.426 & 0.763 \\
V2L w/o $\mathcal{L}_{\mathrm{rele}}$ & 0.404 & 0.987 & 0.906 & 0.908 & 0.481 & 0.779 & 0.546 & 0.930 & 0.833 & 0.852 & 0.443 & 0.787 \\
V2L w/o $\mathcal{L}_{\mathrm{post}}$ & 0.401 & 0.987 & 0.901 & 0.903 & 0.479 & 0.769 & 0.546 & 0.929 & 0.828 & 0.846 & 0.448 & 0.782 \\
V2L w/o $\mathcal{L}_{\mathrm{ib}}$ & 0.395 & 0.987 & 0.894 & 0.896 & 0.473 & 0.756 & 0.539 & 0.929 & 0.823 & 0.842 & 0.439 & 0.776 \\
\rowcolor{MyBlue}V2L & {0.426} & {0.988} & {0.915} & {0.916} & 0.494 & {0.793} & {0.562} & {0.934} & {0.836} & {0.858} & {0.473} & {0.793} \\
\bottomrule[1.1pt]
\end{tabular}}
\vspace{-0.35cm}
\end{table*}

\begin{table*}[t!]
\caption{Ablation results with respect to fusion strategy and missing-index guidance on Corel5k and Pascal07 datasets under 50\% missing views and 50\% missing labels. `w/o' means `without'.}
\label{tab:abl_fusion_index}
\footnotesize
\centering
\resizebox{0.99\textwidth}{!}{
\rowcolors{3}{LightGray}{white}
\begin{tabular}{c|cccccc|cccccc}
\toprule[1.1pt]
\multirow{2}{*}{Method} & \multicolumn{6}{c|}{Corel5k} & \multicolumn{6}{c}{Pascal07} \\
& AP & 1-HL & 1-RL & AUC & 1-OE & 1-Cov & AP & 1-HL & 1-RL & AUC & 1-OE & 1-Cov \\
\midrule
V2L w/ static fusion & 0.416 & 0.987 & 0.905 & 0.908 & {0.498} & 0.783 & 0.551 & 0.931 & 0.831 & 0.853 & 0.450 & 0.788 \\
V2L w/ passive fusion & 0.412 & 0.987 & 0.907 & 0.910 & 0.484 & 0.782 & 0.550 & 0.929 & 0.831 & 0.854 & 0.448 & 0.790 \\
V2L w/o late fusion & 0.421 &0.988 & 0.910& 0.913 &0.493&0.790&0.559&0.934&0.834&0.857&{0.471}&0.790\\
V2L w/o perm. consist. & 0.405 & 0.987 & 0.906 & 0.908 & 0.486 & 0.788 & 0.554 & 0.933 & 0.826 & 0.852 & 0.462 & 0.783 \\
V2L w/o view index & 0.398 & 0.987 & 0.900 & 0.903 & 0.480 & 0.769 & 0.540 & 0.930 & 0.825 & 0.847 & 0.450 & 0.779 \\
V2L w/o label index & 0.403 & 0.987 & 0.906 & 0.909 & 0.468 & 0.781 & 0.532 & 0.928 & 0.833 & 0.853 & 0.435 & 0.789 \\
\rowcolor{MyBlue}V2L & {0.426} & {0.988} & {0.915} & {0.916} & 0.494 & {0.793} & {0.562} & {0.934} & {0.836} & {0.858} & {0.473} & {0.793} \\
\bottomrule[1.1pt]
\end{tabular}}
\vspace{-0.35cm}
\end{table*}
\subsection{Ablation Study}

To evaluate the contribution of each loss component in V2L, we conduct loss function ablation experiments on Corel5k and Pascal07 under 50\% missing views and 50\% missing labels. Specifically, we remove the posterior regularization loss $\mathcal{L}_{\mathrm{post}}$, the information bottleneck loss $\mathcal{L}_{\mathrm{ib}}$, and the relevance regularization $\mathcal{L}_{\mathrm{rele}}$ individually or in combination, and report the results in Table~\ref{tab:abl_loss}. Several observations can be drawn: The complete model achieves the best performance on all metrics across both datasets, showing that the three losses are complementary rather than redundant; Removing any single loss causes a clear performance drop, which verifies that representation alignment, semantic compression, and view-label relevance learning all contribute to the final prediction quality. Among them, removing $\mathcal{L}_{\mathrm{ib}}$ leads to the most obvious degradation in most cases, especially on AP, AUC, and 1-Cov, indicating that the self-supervised information bottleneck design provides the most fundamental support for learning discriminative and semantically sufficient latent representations; When multiple losses are removed simultaneously, the performance declines further, and the worst results are obtained when all three auxiliary losses are discarded, which further confirms the necessity of jointly optimizing these components in the proposed framework.

In addition to the loss functions, we further investigate the effect of the proposed active post-fusion strategy and the prior missing index guidance. The corresponding results are reported in Table~\ref{tab:abl_fusion_index}. To be specific, replacing the proposed instance-wise label-aware fusion with static fusion (All available views are assigned equal weight in late-fusion, ``V2L w/ static fusion'') or passive fusion (All available views are weighted by a group of learnable view-level parameters, ``V2L w/ passive fusion'') leads to degrade overall performance on both datasets. It's easy to find that these simpler variants underperform the complete V2L on the most representative metrics such as AP, AUC, and 1-OE, indicating that fixed or globally learnable view weights are insufficient to capture the complex view-label heterogeneity in iM3C. In contrast, our active perception-based fusion can adapt the contribution of each view to different labels at the instance level, thereby yielding more customized fusion. Furthermore, we completely removed the late-fusion strategy, employing only a combination of mid-level fusion and a classifier (i.e., representation-only fusion version published in \cite{liu2026permutation}, termed ``V2L w/o late fusion''). Experimental results show that if the decisions of each branch are not properly balanced, the hybrid fusion approach may actually lead to worse results due to decision conflicts. 

In addition to the ablation of fusion strategies, removing the missing-view index matrix $\mathbf{W}$ or the missing-label index matrix $\mathbf{R}$ also degrades the model, which confirms the necessity of explicitly incorporating prior missing information into the learning process. In particular, masking the missing-view information causes a more obvious overall drop on both datasets, showing that reliable identification of available views is crucial for both semantic aggregation and adaptive decision fusion. These results further support that the advantage of V2L comes not only from stronger information-centered learning objectives, but also from a more appropriate fusion mechanism and explicit missing-aware modeling.

\subsection{Hyper-Parameter Analysis}
\begin{figure}[t]
	\centering
	\subfloat[Corel5k]{
		\includegraphics[width=0.48\linewidth]{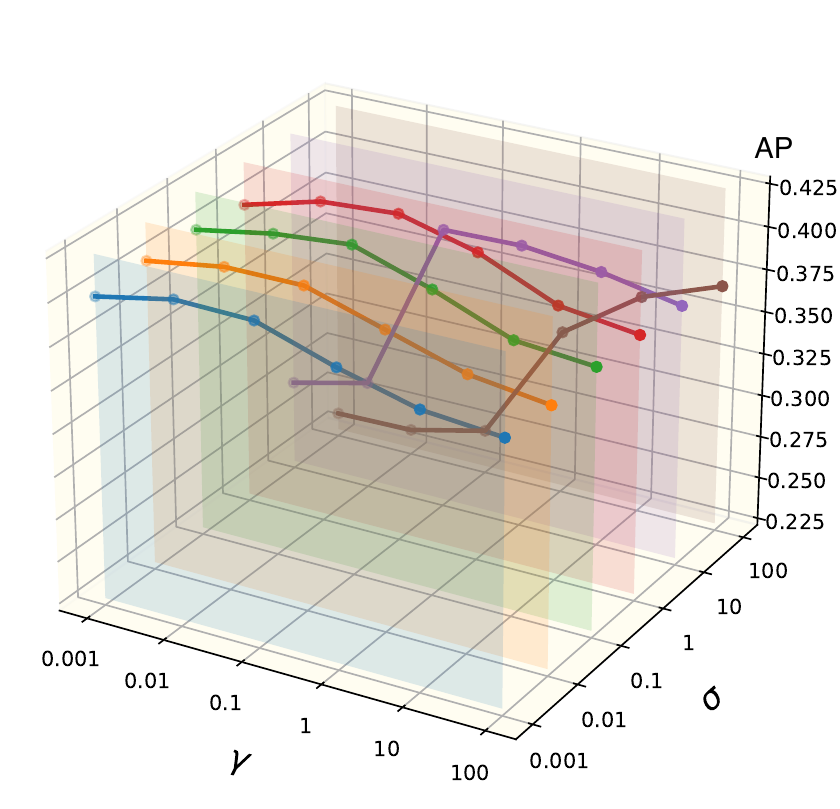}
	}
	\subfloat[Pascal07]{
		
		\includegraphics[width=0.48\linewidth]{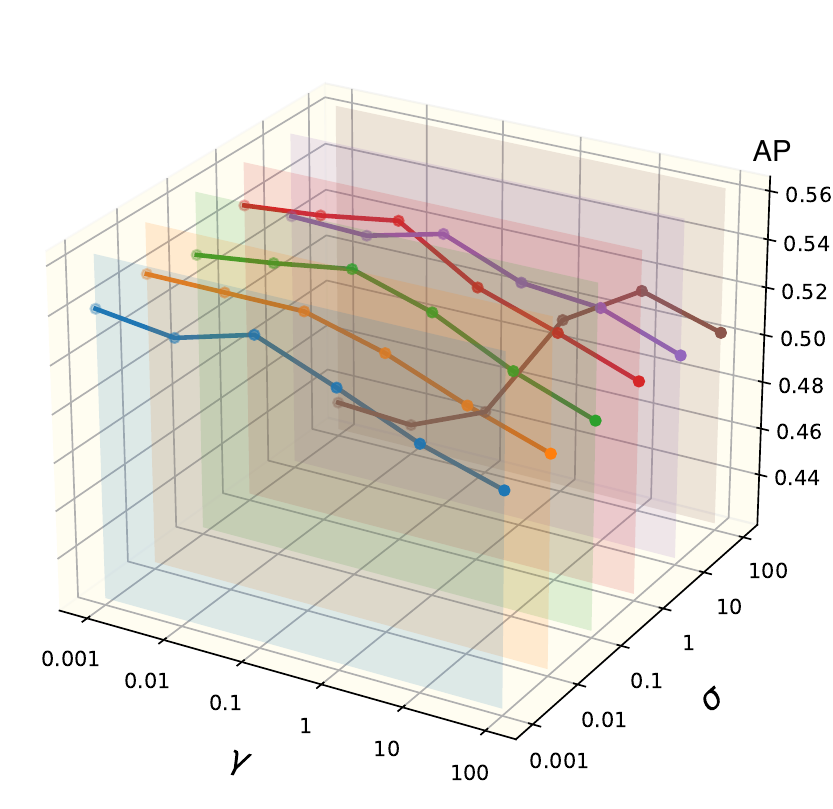}
	}
	\caption{Hyper-parameter sensitivity analysis of $\gamma$ and $\sigma$ on Corel5k and Pascal07 datasets with 50\% missing views and labels.}
	\label{fig.para_alphagamma}
	\vspace{-0.3cm}
\end{figure}

In this subsection, we further analyze the sensitivity of the two trade-off coefficients $\gamma$ and $\sigma$ in the overall optimization. Fig.~\ref{fig.para_alphagamma} reports the performance variations on Corel5k and Pascal07 under different parameter combinations. Overall, the two datasets show a consistent tendency: V2L is relatively stable when both coefficients are chosen in a moderate range, whereas excessively large values of either parameter lead to clear performance degradation. 
More specifically, when $\gamma$ is too small, the cross-view posterior regularization effect is insufficient, so the latent representations cannot fully benefit from the shared semantics among incomplete views. As a result, the model tends to preserve more view-specific disturbance and the final prediction quality is suboptimal. On the contrary, when $\gamma$ becomes too large, the regularization constraint becomes overly strong and suppresses necessary view-specific discriminative cues, which is unfavorable for our view-label heterogeneity modeling. A similar phenomenon can also be observed for $\sigma$: Small or moderate $\sigma$ helps maintain effective information bottleneck-based regularization, while overly large $\sigma$ causes obvious performance drop, especially on Corel5k when $\sigma$ increases to $10$ or $100$. 

According to Fig.~\ref{fig.para_alphagamma}, the best results are generally achieved when $\gamma$ is selected around $10^{-2}\sim10^{-1}$ and $\sigma$ is selected around $10^{-1}\sim10^{0}$. In this region, the model can reach a favorable balance, leading to consistently strong performance on all datasets.

\begin{table*}[t]
    \centering
    \caption{
    Comparison of runtime, parameter count, and peak GPU memory
    consumption on the Corel5k dataset with 70\% training samples.
    Runtime is reported in seconds, parameter count in millions, and
    GPU memory in MB.
    }
    \label{table.time}
    \resizebox{0.95\textwidth}{!}{
    \begin{tabular}{l*{11}{c}}
        \toprule[1.1pt]
        \diagbox{Metric}{Method}
        & CDMM& DM2L& LVSL& iMVWL& NAIM3L& DICNet& DIMC& MSLPP& SIP& QARF& V2L \\\midrule

        Training time (s)& 16.02& 713.37& 63.73& 165.82& 143.63& 313.89& 141.85& 4889.84& 336.11& 134.03& 507.13 \\

        Inference time (s)& 1.73& 0.04& 0.64& 0.02& 0.01& 0.05& 0.04& 0.05& 0.01& 0.02& 0.03 \\

        Parameters (M)& --& --& --& --& --& 187.232& 111.142& 82.897& 34.824& 81.265& 73.377 \\

        Peak GPU memory (MB)& --& --& --& --& --& 2895.40& 2172.65& 1295.59& 1031.64& 1271.72& 1803.68 \\

        \bottomrule[1.1pt]
    \end{tabular}}
    \vspace{-0.5cm}
\end{table*}

{\subsection{Time Cost Study}}

{We evaluate the computational cost of all methods on Corel5k using the same data split, with $70\%$ of the samples for training. Because model training time is highly sensitive to convergence criteria, we measure all methods under their default convergence settings. For single-view methods, we record the total training time summed over all views, and the inference time for a single view. For DICNet, SIP, and PCVE, we conduct 100 epochs for the training phase. As reported in Table~\ref{table.time}, V2L requires $507.13$ seconds for training. Although this is higher than several simpler baselines due to the proposal encoders, posterior regularization, and hybrid fusion architecture, it remains lower than DM2L and MSLPP.}

{V2L requires only $0.03$ seconds for inference because the reconstruction decoders and posterior regularization terms are used only during training. In terms of model resources, V2L contains $73.377$ million trainable parameters and consumes $1803.68$ MB of peak GPU memory. Its parameter count is lower than those of DICNet, DIMC, MSLPP, and QARF, while its memory consumption is lower than those of DICNet and DIMC. Overall, V2L introduces moderate and manageable training overhead while maintaining efficient inference. Parameter count and GPU memory are reported only for neural methods, since these metrics are not directly comparable for conventional optimization-based methods.}

\section{Conclusion}
In this paper, we have presented V2L, a unified framework for incomplete multi-view multi-label classification. From the perspective of multi-view representation learning, we argue that reliable prediction under incomplete observations should be built upon semantically consistent representation learning. Accordingly, we develop a source-perturbation invariant encoding scheme to preserve task-relevant shared information while suppressing view-specific redundancy, so that the learned latent representations remain both informative and consistent across views. From the perspective of view-label heterogeneity modeling, we further emphasize that shared semantics alone are not sufficient for multi-label prediction, since different labels may depend on different views in an instance-dependent manner. Therefore, we introduce an active view-label relevance modeling strategy to characterize label-aware and sample-specific view complementarity beyond static fusion. These two strategies are finally integrated through a hybrid fusion architecture that combines mid-level semantic fusion and late-stage adaptive decision fusion within a unified model. The former provides stable shared semantic evidence for incomplete multi-view representation learning, while the latter preserves complementary view-specific cues for label prediction. Extensive experiments and analyses on benchmark datasets verify that this design yields both superior predictive performance and interpretable view-label relevance patterns. 

The proposed strategy may also be extended to other incomplete multi-view or multimodal tasks: the source-perturbation invariant encoding can provide robust shared representations, while the active relevance mechanism can model target-dependent contributions from different information sources \cite{li2018survey,yan2025incomplete,yin2026clustering}. Extending V2L to tasks such as multimodal diagnosis, multi-sensor recognition, and multimedia understanding by designing task-specific relevance supervision and prediction objectives is an important direction for future work.

\bibliographystyle{IEEEtran}
\bibliography{pami26}

\vspace{-0.7cm}
\begin{IEEEbiography}[{\includegraphics[width=1in,clip,keepaspectratio]{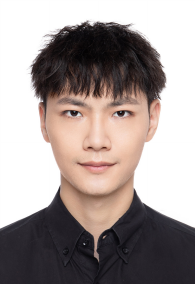}}]{Chengliang Liu} received the PhD degree in computer science from Harbin Institute of Technology, Shenzhen, China, in 2024; the M.S. degree in computer science from Huazhong University of Science and Technology, Wuhan, China, in 2020; and the B.S. degree in computer science from Jilin University, Changchun, China, in 2018. He is currently a postdoctoral fellow at the University of Macau, Macau. He serves as an Associate Editor of \textit{Pattern Recognition}, Guest Editor of \textit{Information Fusion}, and SPC of IJCAI and AAAI. His research interests include machine learning and computer vision, especially multimodal learning. Dr. Liu was honored with the Distinguished Paper Award at AAAI 2023, and he received the Nomination Award for CSIG Doctoral Dissertation Incentive Program and Harbin Institute of Technology's Outstanding Dissertation Award. His homepage: \url{https://justsmart.github.io}.
\end{IEEEbiography}
\vspace{-0.7cm}
\begin{IEEEbiography}[{\includegraphics[width=1in,clip,keepaspectratio]{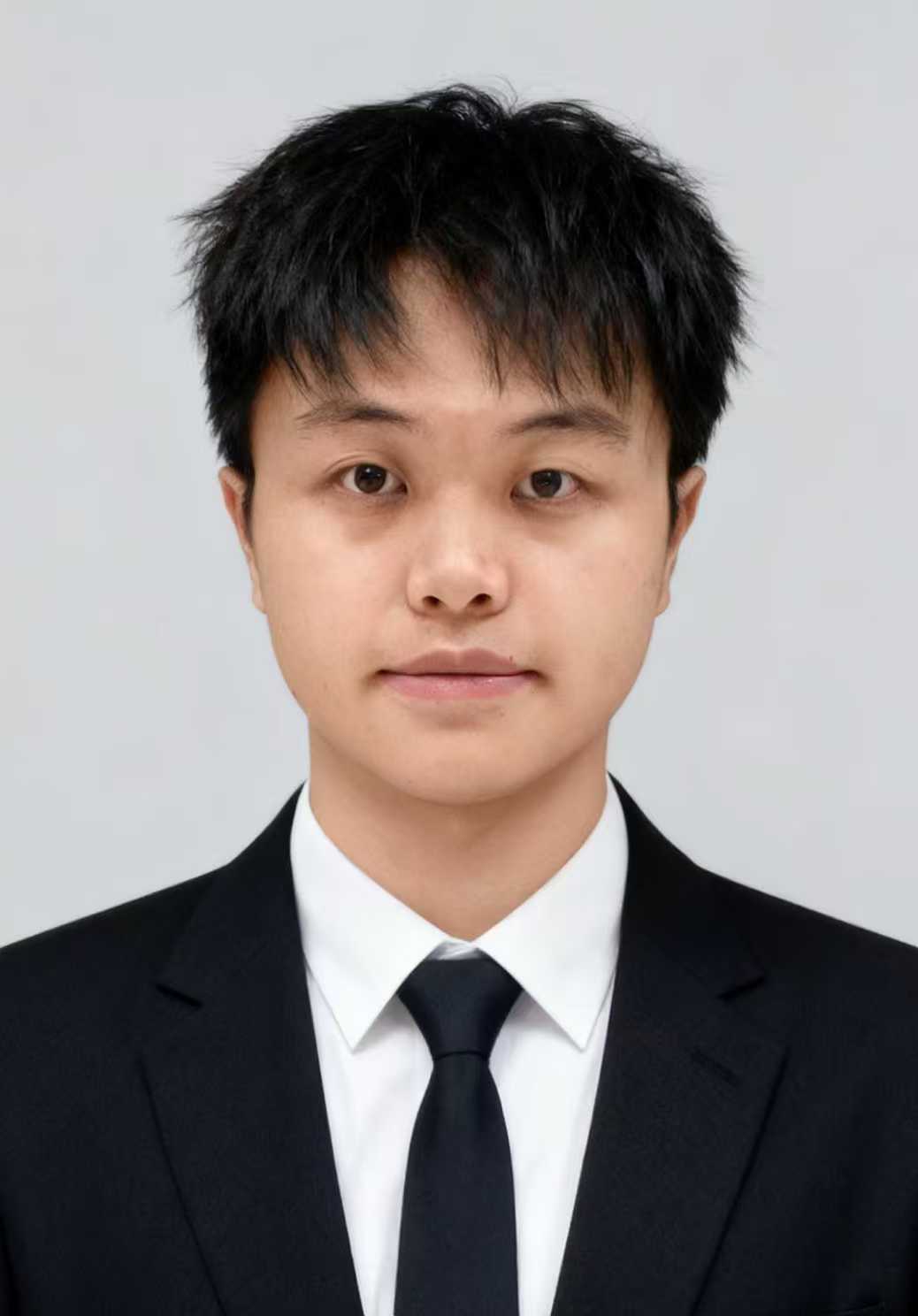}}]{Bo Li}
received his B.Eng. degree in Robotics Engineering from Beijing Information Science and Technology University in 2021. He received his M.Eng. degree in Electronic Information from Beijing University of Technology in 2024, and is currently a Ph.D. student in Computer Science at the University of Macau. His research lies at the intersection of AI for biomedicine and agent systems, with interests in biomedical image analysis, image-based phenotypic drug discovery, virtual cell, and biomedical knowledge retrieval.
\end{IEEEbiography}
\vspace{-0.5cm}
\begin{IEEEbiography}[{\includegraphics[width=1in,clip,keepaspectratio]{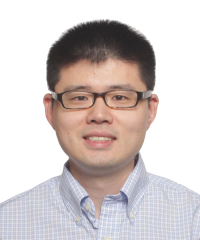}}]{Bob Zhang}
(Senior Member, IEEE) received the B.A. degree in computer science from York University, Toronto, ON, Canada, in 2006, the M.A.Sc. degree in information systems security from Concordia University, Montreal, QC, Canada, in 2007, and the Ph.D. degree in electrical and computer engineering from the University of Waterloo, Waterloo, ON, Canada, in 2011. He is currently an Associate Professor and \textbf{Associate Head} of the Department of Artificial Intelligence, University of Macau, Macau. His research interests include biometrics, pattern recognition, and image processing. He is a Technical Committee Member of the IEEE SMC Society and Associate Editors of IEEE TIP, IEEE TSMC: System, IEEE TNNLS, the IEEE OJ-CS, and Artificial Intelligence Review. 
\end{IEEEbiography}
\vspace{-0.5cm}
\begin{IEEEbiography}[{\includegraphics[width=1in,clip,keepaspectratio]{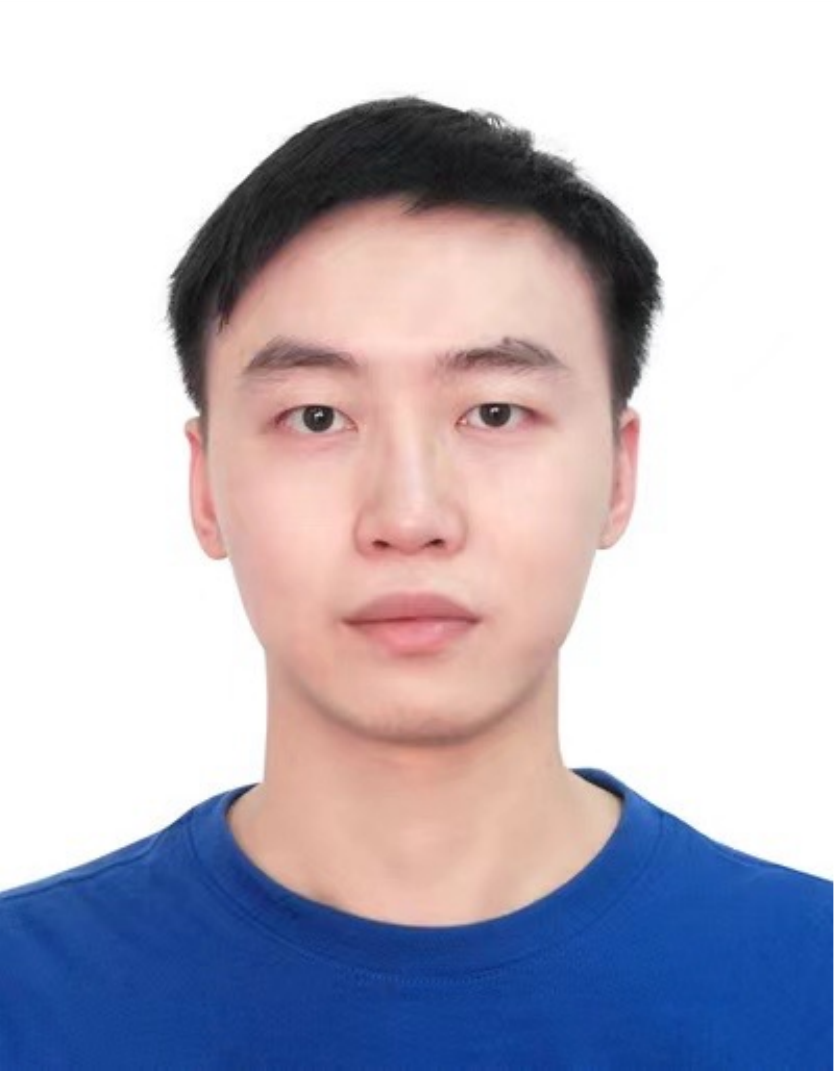}}]
{Yanghao Zhou} (Student Member, IEEE) received the Bachelor’s Degree from Jilin University, China, in 2018 and received the Master's Degree from The University of Sheffield, in 2020. He is currently pursuing a PhD degree in the School of Computer Science and Technology, Beijing Institute of Technology, China. His research interests are in Robust Multi-Modal Learning and its Application, Multimodal Large Language Models, and AIGC.
\end{IEEEbiography}
\vspace{-0.5cm}
\begin{IEEEbiography}[{\includegraphics[width=1in,clip,keepaspectratio]{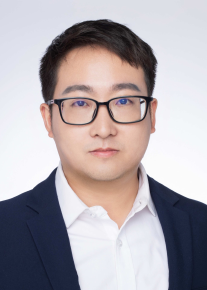}}]{Jie Wen} (Senior Member, IEEE) received the Ph.D. degree in Computer Science and Technology at Harbin Institute of Technology, Shenzhen in 2019. He is currently a Professor at the School of Computer Science and Technology, Harbin Institute of Technology, Shenzhen. His research interests include image and video enhancement, pattern recognition, and machine learning. He has authored or co-authored more than 100 technical papers at prestigious international journals and conferences, including the TNNLS, TIP, TCYB, NeurIPS, ICML, CVPR, AAAI, IJCAI, ACM MM, etc. He serves as an \textbf{Associate Editor} of \textit{IEEE Transactions on Pattern Analysis and Machine Intelligence}, \textit{IEEE Transactions on Image Processing}, \textit{IEEE Transactions on Information Forensics and Security},  \textit{Pattern Recognition}, and \textit{International Journal of Image and Graphics}, an \textbf{Area Editor} of \textit{Information Fusion}. He is on the \textbf{Young Editorial Board} of \textit{CAAI Transactions on Intelligence Technology}. He also served as the \textbf{Area Chair} of \textit{NeurIPS}, \textit{ICLR}, \textit{ACM MM}, and \textit{ICML}, as well as the SPC of \textit{AAAI} and \textit{IJCAI}. 
\end{IEEEbiography}
\vspace{-0.8cm}
\begin{IEEEbiography}[{\includegraphics[width=1in,clip]{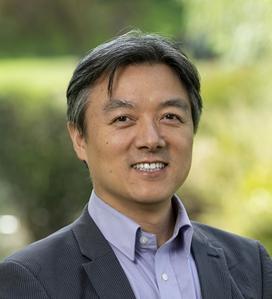}}]
{Wenwu Wang} (Fellow, IEEE) was born in Anhui, China. He received the B.Sc., M.E., and the Ph.D. degrees, all in the field of automation, from Harbin Engineering University, China, in 1997, 2000, and 2002, respectively. He then worked with King’s College London, Cardiff University, Tao Group Ltd. (now Antix Labs Ltd.), and Creative Labs, before joining University of Surrey, U.K., in May 2007, where he is currently a Professor in Signal Processing and Machine Learning, and an Associate Head in External Engagement, School of Computer Science and Electronic Engineering, University of Surrey, UK. He is also a Principal AI Fellow at the Surrey Institute for People Centred Artificial Intelligence. His current research interests include signal processing, machine learning and perception, artificial intelligence, and statistical anomaly detection. He has (co)-authored over 400 papers in these areas. His works have been recognized with various awards, including the Meta Distinguished Faculty Award (2026), Audio Engineering Society Best Technical Paper Award (2025), IEEE Signal Processing Society Young Author Best Paper Award (2022), DCASE Judge’s Award (2020, 2023, and 2024), DCASE Reproducible System Award (2019 and 2020), and LVA/ICA Best Student Paper Award (2018). He is an IEEE Fellow. He has been a keynote or plenary speaker at 30+ international conferences and workshops.
\end{IEEEbiography}

\vfill

\end{document}